\documentclass[10pt,twocolumn,letterpaper]{article}

\usepackage{cvpr}              

\usepackage{graphicx}
\usepackage{amsmath}
\usepackage{amssymb}
\usepackage{booktabs}
\usepackage{import}
\usepackage{float}
\usepackage{adjustbox}
\usepackage{algorithm}
\usepackage{tabularx}
\usepackage{multirow}
\usepackage{color}
\usepackage{colortbl}
\usepackage{xcolor}
\usepackage{array, makecell}
\usepackage[noend]{algpseudocode}
\usepackage{wrapfig}
\usepackage{titling}

\usepackage[pagebackref,breaklinks,colorlinks]{hyperref}

\usepackage[capitalize]{cleveref}

\newcommand{\bandc}{\cellcolor{blue!6}}

\crefname{section}{Sec.}{Secs.}
\Crefname{section}{Section}{Sections}
\Crefname{table}{Table}{Tables}
\crefname{table}{Tab.}{Tabs.}

\begin{document}

\title{Variation of Gender Biases in Visual Recognition Models \\ Before and After Finetuning }
\author{%
\begin{tabular}{c} Jaspreet Ranjit \\ University of Southern California \\ {\tt\small jranjit@usc.edu} \\ \\
Baishakhi Ray \\ Columbia University \\ {\tt\small rayb@cs.columbia.edu} \end{tabular} \and
\begin{tabular}{c} Tianlu Wang \\ Meta AI \\ {\tt\small tianluwang@meta.com}  \\ \\
Vicente Ordonez \\ Rice University \\ {\tt\small vicenteor@rice.edu}\\
\end{tabular} }

\date{}
\maketitle

\begin{abstract}
    We introduce a framework to measure how biases change before and after fine-tuning a large scale visual recognition model for a downstream task. Deep learning models trained on increasing amounts of data are known to encode  societal biases. Many computer vision systems today rely on models typically pretrained on large scale datasets. While bias mitigation techniques have been developed for tuning models for downstream tasks, it is currently unclear what are the effects of biases already encoded in a pretrained model. Our framework incorporates sets of canonical images representing individual and pairs of concepts to highlight changes in biases for an array of off-the-shelf pretrained models across model sizes, dataset sizes, and training objectives. Through our analyses, we find that (1) supervised models trained on datasets such as ImageNet-21k are more likely to retain their pretraining biases regardless of the target dataset compared to self-supervised models. We also find that (2) models finetuned on larger scale datasets are more likely to introduce new  biased associations. Our results also suggest that (3)  biases can transfer to finetuned models and the finetuning objective and dataset can impact the extent of transferred biases.

\end{abstract}

\section{Introduction}
\label{sec:intro}
Most visual recognition models today are not trained from scratch but instead rely on some model pretrained on a large scale dataset that has been {\em finetuned} to perform well on a target task. These pretrained models are widely available and can be easily adopted and repurposed with minimal effort. However, adopting a third-party component in any system introduces some risks, as these models might introduce behaviors or associations that are unknown to the adopter.  Our work proposes a framework to audit and analyze these models regarding biases with respect to sensitive visual attributes and the potential inadvertent effects of adopting them for a target task. 

Visual recognition models are known to encode societal biases~\cite{hendricks2018women,zhao2017men,buolamwini2018gender,wilson2019predictive,singh2020don}. These biases are especially problematic when they are related to protected or sensitive attributes such as perceived gender, sex, or ethnicity. In our work, we rely on reference sets of images to represent concepts and characterize the associations these models encode in their latent representations. For instance, consider a group of images representing the label \texttt{women}, and another group representing the label \texttt{men} as the perceived gender.
A model can be characterized as containing bias with respect to perceived gender if a group of images labeled \texttt{surfboard} is consistently mapped in the same embedding space 
as one of the two former groups (\texttt{women} or \texttt{men}). 
These associations might occur in the latent representation space even if the model was not trained to predict these concepts.
Our approach is widely applicable as we measure associations across reference images without requiring the model to predict labels. Moreover, our work contributes to prior studies by exploring the dynamic of these associations before and after finetuning.

We empirically evaluate an extensive number of pretrained models across model capacities, datasets, and training objectives (supervised vs self-supervised) and analzye how biases vary after model fine-tuning. 

    Our work introduces several contributions to understand the problem of how biases change in the fine-tuning process. First, we curate reference analysis sets~\cite{steed2021image} for quantifying gender biases in latent visual representations based on the COCO~\cite{coco} and OpenImages~\cite{openimages} datasets. Beyond measuring the associations in the latent space with respect to {\em analysis sets}, our work proposes to further study how these associations vary after finetuning to measure the persistency of these associations. 
    Our findings show that our measure serves as a valid indicator of bias retention and provides insights regarding model behavior in the latent space. 
   Furthermore, we analyze factors contributing to bias transfer through extensive experiments. We particularly observe that models trained with supervised objectives on large scale datasets are more likely to retain their pretraining biases after finetuning than models trained with self-supervised objectives. We also observe that models trained on larger scale target datasets are more likely to exhibit new gender biased associations after finetuning compared to those trained on smaller scale target datasets. We particularly define a Bias Transfer Score (BTS) based on Spearman's correlation coefficient to capture the dynamics of intra-class and inter-class similarities induced by the latent representations of visual recognition models before and after finetuning.

We expect that as new models are released, the numbers presented as measures for intra-class and inter-class variation as proposed in our work will serve as reference for future adopters of such models.

\begin{figure*}[t]
     \centering
         \includegraphics[width=\textwidth]{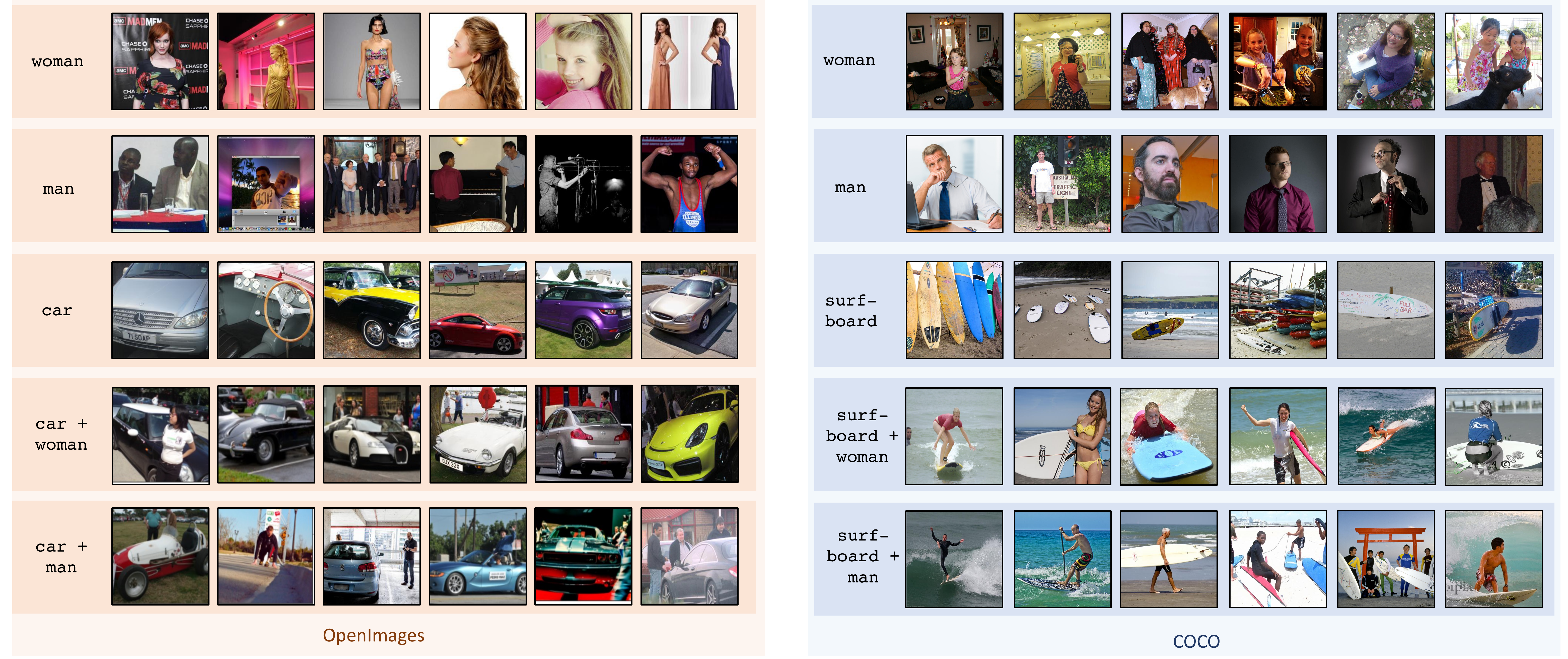}
     \caption{Sample images from our analysis sets to analyze correlations between images of people and everyday objects based on the Open Images and COCO datasets. Given this set of images, we can probe whether visual representations from a pretrained neural network embed the representations for images in the set \texttt{man} and \texttt{car} closer together relative to \texttt{woman} and \texttt{car}.
              OpenImages contains annotations for the perceived gender of people, while COCO contains captions that often refer to the perceived gender of people in pictures. Images are manually selected so that each analysis set contains images representing a single object and images that represent a single object co-occurring with a person with an associated perceived gender.}%
     \label{fig:analysis_set_coco}
 \end{figure*}

\section{Related Work} 
\label{sec:relatedworks}

Our work is related to a growing body of work studying biases in visual recognition~\cite{hendricks2018women,sadeghi2020imparting,wang2019balanced,choi2019can,wang2020towards,thong2021feature,singh2020don}. While some works focus on bias mitigation, our work is more related to those measuring and understanding biases~\cite{zhao2017men,buolamwini2018gender,steed2021image,de2019does,wang2020revise,balakrishnan2020towards,tian2020testing,miceli2021documenting,serna2021insidebias,salman2022does}. 

\vspace{0.01in}
\noindent\textbf{The Issue of Bias:} The phenomena of dataset bias has been a concern in computer vision for some time, where models trained on a specific set of categories fail to transfer to other datasets~\cite{torralba2011unbiased}. 
More recently, the phenomena of bias with respect to protected or sensitive attributes has received attention. For instance, facial recognition systems have been found to be biased in their predictions for different demographic groups~\cite{buolamwini2018gender}.
Biases with respect to sensitive attributes have also been found in models targeting more general computer vision tasks such as object classification~\cite{zhao2021captionbias}, image captioning~\cite{hendricks2018women}, and object detection~\cite{wilson2019predictive}. Understanding more carefully how biases are introduced in computer vision models, and how to mitigate them are ongoing questions. Our work contributes to these questions with a method to measure how re-using models that have been pretrained on large scale datasets might introduce biases. 
\vspace{0.01in}
\noindent\textbf{Measuring Bias at the Prediction Level:} Our work is complementary to others such as REVISE~\cite{wang2020revise} which targets the identification of biases in datasets and VisCUIT~\cite{lee2022viscuit} which aims to highlight activated neurons that lead to biased outputs. More recently, Meister et al.~\cite{meister2022gender} investigates gender artifacts or visual cues that correlate with gender in the COCO and OpenImages datasets. Our work, assumes that as large scale pretrained models are released, training datasets will often not be publicly available for scrutiny as is the case for the CLIP model~\cite{clip}. Moreover, as studies that focus on bias mitigation demonstrate, models can often be trained to comparable accuracy while learning qualitatively different biases on the same dataset~\cite{hendricks2018women,wang2019balanced,sadeghi2020imparting,choi2019can,wang2020towards}. Our work reveals how biases propagate through finetuning to offer practical guidance to both practitioners and researchers.

\vspace{0.01in}
\noindent\textbf{Measuring Bias at the Feature Level:} There is a long literature in machine learning in designing methods to quantify bias through the adoption of fairness criteria such as equality of odds, or equality of opportunity~\cite{hardt2016equality,heidari2019moral}. In computer vision, there have been some efforts in adopting similar fairness criteria to evaluate model predictions~\cite{zhao2017men,wang2020towards,buolamwini2018gender}. However, our work is more concerned with {\em representational bias}, where we aim to measure the biases in the model's internal representations and not its predictions. In the field of natural language processing, Bolukbasi~et~al~\cite{bolukbasi2016man} studied biases in the representations of words in English by conducting association tests. More recently, Steed and Caliskan~\cite{steed2021image} adopted association tests for measuring biases in image representations from self-supervised models through the use of {\em analysis sets} for representing a concept. Our work goes beyond and measures how bias changes from pre-training to fine-tuning. More related to ours, the recent work of Salman et al.~\cite{salman2022does} is the first to show evidence of bias transfer from pre-training models using both synthetic and real data. Our work further proposes a Bias Transfer Score (BTS) for quantifying the variation in biases before and after fine-tuning a model.

\section{Background} 

A common approach in computer vision involves re-using a model that has been pre-trained on a large scale dataset, a {\em pretrained model}, and adapting it to a target task through fine-tuning its model parameters on a smaller scale dataset, thus obtaining a {\em finetuned model}. There are several models in computer vision that have been widely adopted as pretrained models such as ResNets trained on ImageNet~\cite{resnet} and more recently, many works use vision transformers trained on weakly annotated data (i.e.~CLIP~\cite{clip}). Adopters of these models might not have access to the computational resources or data needed to replicate the training of these models, and given their wide availability in the form of pretrained model parameters, they are commonly used as {\em blackboxes}. This introduces some risks, as these models contain biases that might be propagated to target tasks with unintended effects. Our work, provides a framework to comprehensively evaluate a diverse set of such models across a range of criteria, including ResNets of different capacities trained on Imagenet-1k~\cite{russakovsky2015imagenet}, a ResNet trained on Imagenet-21k~\cite{deng2009imagenet}, unsupervised ResNets trained on Imagenet-1k~\cite{he2019moco,chen2020simple}, and a vision transformer model trained on web data~\cite{clip}. Our goal is not only to measure what associations these models contain but also how these associations change after the model is finetuned on target tasks.

\section{Methodology}

We analyze bias by computing similarities between image representations. 
Similarity scores and analogy tests have been used to assess how word representations correlate with each other in the embedding space~\cite{bolukbasi2016man}. 
 We accomplish this in two ways: intra-class and inter-class similarity. We measure the model's variation in the encoding of a single class of images using intra-class similarity to better understand how different models represent image instances of the same class. 
Alternatively, we also measure the similarity between encodings of two different classes using inter-class similarity to examine biased correlations between a perceived gender group and a class. 

Our work investigates the biases of a model with respect to a given class of images by measuring intra-class and inter-class similarities at the feature representation level. Given an {\em analysis set} as defined in Equation~\ref{eqn:analysiset}, where $\mathcal{C}$ is the set of all classes, and $k$ is the number of examples in class $c$,
\begin{equation}
\label{eqn:analysiset}
A = \{x_{i}^{c} | c \in \mathcal{C}, i \in 1 \ldots k\},
\end{equation}
with a given model $M(*)$, for each class $c$, our framework first obtains a set of intermediate representations $\mathcal{Z}^c= M(x_{i}^{c})$ as defined in Equation \ref{eqn:features} where $d$ is the feature embedding size for model $M(*)$:
\begin{equation}
\label{eqn:features}
\mathcal{Z}^c = \{z_i| z_i\in\mathbb{R}^d, i \in \ldots k\}.
\end{equation}
Given these preliminaries, then we proceed to define our measures for intra-class and inter-class similarity.

\subsection{Intra-Class Similarity}
This measure aims to capture on average how much the model clusters images together in its representation space for a given class $c$. We expect that for some class of images with homogeneous looking objects such as \texttt{stop-signs} that are mostly red and exhibit a similar shape, most models will on average represent them closer in feature space than for other classes that exhibit more intra-class variation such as \texttt{chairs} which can come in different shapes and colors. For intra-class similarity, we randomly split the feature set $\mathcal{Z}^c$ into two groups of equal size $\mathcal{Z}_1^c$ and $\mathcal{Z}_2^c$. We then randomly take two examples $z_1^c$ and $z_2^c$ from each group and compute their similarity score $s = \phi(z_1^c, z_2^c)$. We repeat this random sampling process for $m$ iterations and compute an average similarity score $\mu_c$ for a given class $c$. As defined in Equation \ref{eqn:intra}, $\mu_c$ provides us with a sense of intra-class variation in an analysis set and can further serve as a proxy to identify biases in a model. 
When compared relatively to other classes, intra-class similarity can help us understand how well a model has learned to represent a class and we can better observe the changes in the feature representations of a particular class after finetuning.

Algorithm \ref{alg:intraclass} summarizes how to compute intra-class similarity. 
\begin{equation}
\label{eqn:intra}
\mu_{c} = \frac{1}{m} * \sum_{j=1}^{m}\phi(z_1^c, z_2^c); \hspace{0.15cm} z_1^c \in \mathcal{Z}_1^c, z_2^c\in\mathcal{Z}_2^c
\end{equation}

\begin{algorithm}[t]
\captionsetup{font=small} 
\small
\caption{{Intra-Class Similarity}}\label{alg:intraclass}  \textbf{Input:} {Analysis Set: \\ \hspace*{\algorithmicindent}\hspace*{\algorithmicindent}\hspace{0.13cm}$\{x_{i}^{c} \mid {c \in \mathcal{C}}; i\in 1 ... k\}$; \\
\hspace*{\algorithmicindent}\hspace*{\algorithmicindent}\hspace{0.13cm}Model: $M(*)$; \\
\hspace*{\algorithmicindent}\hspace*{\algorithmicindent}\hspace{0.13cm}Number of iterations: m}\\
\textbf{Output}: {Intra-class similarity for class $p$: $\mu_p$}
\begin{algorithmic}[1]
\State $\mathcal{Z}^p = M(\{x_{i}^{p}\})$
\For{j = 1 to m}
\State ${\hat{\mathcal{Z}^{p}}}$ = permute($\mathcal{Z}^p$)
\State ${\hat{\mathcal{Z}_1^p}}$,${\hat{\mathcal{Z}_2^p}}$ = ${\hat{\mathcal{Z}^{p}}[:k/2]}$, ${\hat{\mathcal{Z}^{p}}[k/2:]}$
\State $z_1^p$ $\leftarrow$ RandomSample(${\hat{\mathcal{Z}_1^p}}$)
\State $z_2^p$ $\leftarrow$ RandomSample(${\hat{\mathcal{Z}_2^p}}$)
\State $s_j$ = $\phi(z_1^p, z_2^p)$
\EndFor
\State $\mu_{p} = \frac{1}{m} * \sum_{j=0}^{m} s_j$
\end{algorithmic}
\end{algorithm}
\begin{algorithm}[t]
\captionsetup{font=small} 
\small
\caption{\small{Inter-Class Similarity}}\label{alg:interclass}
\textbf{Input:}{Analysis Set: \\
\hspace*{\algorithmicindent}\hspace*{\algorithmicindent}\hspace{0.13cm}$\{x_{i}^{c} \mid {c \in \mathcal{C}}; i\in 1 ... k\}$; \\
\hspace*{\algorithmicindent}\hspace*{\algorithmicindent}\hspace{0.13cm}Model: $M(*)$; \\
\hspace*{\algorithmicindent}\hspace*{\algorithmicindent}\hspace{0.13cm}Number of iterations: m}\\
\textbf{Output:} {Intre-class similarity between class $p$ and $q$: $\mu_{p, q}$}
\begin{algorithmic}[1]
\vspace{0.12cm}
\State $\mathcal{Z}^{p} = M(\{x_{i}^{p}\})$ 
\State $\mathcal{Z}^{q} = M(\{x_{i}^{q}\})$
\For{ j $=$ 0 to m}
\State $z^{p}$ $\leftarrow$ RandomSample(${{\mathcal{Z}^{p}}}$)
\State $z^{q}$ $\leftarrow$ RandomSample(${{\mathcal{Z}^{q}}}$)
\State $s_j$ = $\phi(z^{p}, z^{q})$
\EndFor
\State $\mu_{p,q} = \frac{1}{m} * \sum_{j=1}^{m} s_j$

\end{algorithmic}
\end{algorithm}

\subsection{Inter-Class Similarity}
This measure aims to capture associations between a pair of classes. For instance, previous work found that \texttt{man} and \texttt{skateboard} were two classes that exhibited a strong relationship~\cite{hendricks2018women}.  Given two classes $c_p, c_q \in \mathcal{C}$, and given sets of features generated from a model $M(*)$ for two classes $\mathcal{Z}^{p}$ and $\mathcal{Z}^{q}$, we randomly pick $z^p$ and $z^q$ and compute their similarity $s = \phi(z^{p}, z^{q})$.
We repeat this random sampling for $m$ iterations to get $m$ values representing the similarity between two classes $c_p$ and $c_q$ in the latent space (Algorithm~\ref{alg:interclass}). We take the average of these $m$ values to get $\mu_{p,q}$ which represents the average similarity between two classes, as defined in Equation \ref{eqn:inter}. 

Using inter-class similarity, we can calculate the association between any target class and a class representing protected attributes (e.g. gender). 
\begin{equation}
\label{eqn:inter}
\mu_{p, q} = \frac{1}{m} * \sum_{j=1}^{m}\phi(z^{p}, z^{q}); \hspace{0.15cm} z^{p} \in \mathcal{Z}^{p}, z^{q}\in\mathcal{Z}^{q}
\end{equation}

We use cosine similarity as our choice for $\phi$ since it is a simple metric that does not have any parameters or depend on any model, and is commonly used for image retrieval.

\subsection{Bias Transfer Score (BTS): Definition}

We evaluate model associations with respect to intra-class and inter-class similarities in two phases: \textit{pretraining} and \textit{finetuning}. In the \textit{pretraining} phase, we use an off-the-shelf version of the model that has been pretrained on a large scale image dataset, and we extract features from hidden layers to calculate the intra-class and inter-class similarities. In the \textit{finetuning} phase, we finetune pretrained models on a target task and extract features from the finetuned model. Our goal is to examine how biases change after finetuning.  Given pretrained model $M_1(*)$ and fine-tuned model $M_2(*)$, we compute the intra-class similarities for all classes in $\mathcal{C}$ so that we obtain $\{\mu^{M_1}_c|c\in\mathcal{C}\}$ and $\{\mu^{M_2}_c|c\in\mathcal{C}\}$. We then define our Bias Transfer Score~(BTS) as the Spearman's coefficient $r_{\texttt{BTS}}$ between these two sets of measurements as follows:
\begin{equation}
\label{eqn:spearman}
r_{\texttt{BTS}} = \rho_{R(\{\mu^{M_1}_c|c\in\mathcal{C}\}), R(\{\mu^{M_2}_c|c\in\mathcal{C}\})},
\end{equation} 
the Spearman's coefficient shows us if two variables are monotonically related even if their relationship is not linear. This can also be applied to compute the $r_{\texttt{BTS}}$ score for inter-class similarities by considering the sets $\{\mu^{M_1}_{c_1,c_2}|c_1,c_2\in\mathcal{D}\}$ and $\{\mu^{M_2}_{c_1,c_2}|c_1,c_2\in\mathcal{D}\}$, where $\mathcal{D} \subset \mathcal{C}\times\mathcal{C}$ is a subset of category pairs that contain a target object class and a class representing a protected attribute. Quantitatively, this allows us to compare a model's bias relationships at the pretraining and finetuning stages where a lower $r_{\texttt{BTS}}$ score implies the biases underwent significant changes after finetuning. By combining the use of inter-class and intra-class similarities and Spearman coefficients to measure bias transfer across models with different training objectives, architectures and target datasets, we are able to observe the extent of biased correlations at a finer granularity than previous work.

\section{Data: Gender Analysis Sets}
\label{sec:analysissets}

We choose to analyze gender as our protected attribute since this is generally recognized as a universal attribute that can be applied to all humans and its biases have been studied and recognized as significant in the context of vision models~\cite{zhao2017men,hendricks2018women,wang2019balanced,sadeghi2020imparting,thong2021feature,wang2021directional,zhao2021captionbias,stone2022epistemic,Hirota_2022_CVPR}. One limitation on using gender as a protected attribute is that current datasets do not contain self-reported gender annotations and thus use the perceived gender of people in pictures as a proxy variable. Another consideration is that these type of annotations mostly reduce the labels to men and women, thus potentially ignoring non-binary gender expressions. While these could be problematic, in our work, this is mitigated by the fact that we do not aim to predict gender from pictures but focus instead on just analyzing how gender expression through pictures affect the associations that visual recognition models make with respect to everyday objects.

Following prior work in this area, we collect analysis sets for a set of object classes that are not defined by the gender of people using them or co-occurring with them~\cite{bolukbasi2016man,steed2021image}.

For example, we should not expect images of the class \texttt{kitchen} to be more associated with women than men as this can potentially reinforce a societal stereotype of women as homemakers.  
The images selected for each analysis set are chosen so that each object class is depicted in isolation or co-occurred with \texttt{man} or \texttt{woman}. For example, for a given object class \texttt{surfboard}, the analysis set include images depicting \texttt{surfboard} in isolation, and \texttt{surfboard+man} and \texttt{surfboard+woman} which include images containing \texttt{man/woman} and the object, and ideally no other confounding objects as shown in Figure \ref{fig:analysis_set_coco}. It is necessary to include images that have objects co-occur with gender as the features for objects depicted in isolation can be orthogonal in the embedding space thus the similarity between them cannot be meaningfully captured using inter-class similarity. To better capture more subtle variations in biases before and after finetuning, we intentionally include classes that contain the object co-occurring with the gender.  
We also ensure that the number of images in each analysis set does not differ by a significant amount. Furthermore, to better capture biases introduced during the pretraining phase, there is no significant overlap between the categories chosen in the analysis sets and datasets typically used for pretraining. 
For the purpose of this study, we collect analysis sets from the COCO and Open Images dataset, examples can be found in Figure~\ref{fig:analysis_set_coco}.
Next, we summarize details about our curation process:

\vspace{0.02in}
\noindent\textbf{COCO.} This dataset has over 200k labeled images with 80 object categories. 
We used the object annotations for the images to extract the analysis set. We want to examine gender with respect to the following object categories: \texttt{car}, \texttt{refrigerator}, and \texttt{surfboard}. We also include a \texttt{random} category which consists of images randomly selected from COCO to serve as a reference. Because COCO does not have explicit \texttt{man} and \texttt{woman} annotations, we filter images that contain \texttt{person} and the objects of interest, and then manually select images that best represent each target as it is typically perceived. For example, we ensure the \texttt{surfboard} category only contains images in the context of the \texttt{beach}. The COCO analysis set contains $206$ images across $13$ classes. To the best of our ability, we also ensure that images in a single class have similar backgrounds to limit variability. 

\vspace{0.02in}
\noindent\textbf{Open Images.} This is a larger but noisier dataset that includes over 9 million images with object annotations.
We select the following categories: \texttt{car}, \texttt{sport}, \texttt{equipment}, \texttt{fashion-accessory}, and  \texttt{mammal}. Similarly, we have a \texttt{random} category as a reference.
Open Images contains gender annotations, so the reference analysis sets for \texttt{man} and \texttt{woman} are more straightforward to collect. We ensured that each class in the analysis set did not contain overlapping images with other classes, however, unlike the COCO analysis set, we did not manually inspect this analysis set and as a result, it contains higher variability and greater diversity within a single class. The Open Images analysis set contains $1925$ images across $16$ classes. Our supplementary material contains detailed statistics of the analysis sets. 

\section{Experimental Setup}

\subsection{Models}
\label{sec:pretrained_models}
Using our framework, we evaluate the following off-the-shelf pretrained models: 

        \textbf{ResNet18}: This model introduced by He~et~al~\cite{resnet}, consists of 18-layers with skip connections and batch normalization. We refer to ResNet18 here as, the widely used version of this model that is trained on the Imagenet-1k classification challenge~\cite{russakovsky2015imagenet}. 

        \textbf{ResNet50}: Similar to ResNet18, but this model has 50 layers~\cite{resnet}. We refer to ResNet50 here as specifically the model trained on the Imagenet-1k classification challenge~\cite{russakovsky2015imagenet}. 
        \textbf{BiT-M-R50x1}: This model also leverages the ResNet50 model architecture but is trained on the full Imagenet dataset containing 21k classes~\cite{kolesnikov2020big}. The original full Imagenet-21k dataset might not be available~\cite{yang2020towards}, checkpoints for this model are publicly available.
        \textbf{CLIP-ViT/B32}: The CLIP model~\cite{clip} consists of a vision transformer model (ViT~\cite{dosovitskiy2021an}) that is trained together with a text transformer model using a contrastive objective over approximately 400 million image and text pairs crawled from the web. We adopt CLIP with a ViT/B-32 transformer.

        \textbf{MoCo ResNet50}: This model leverages ResNet50 as the model architecture and the Imagenet-1k dataset but it is trained in a self-supervised fashion without using label annotations, instead optimizing a contrastive learning objective~\cite{he2019moco}.

        \textbf{SimCLR ResNet50}: This model also leverages both ResNet50 as the architecture and Imagenet-1k as the training set and like the previous model, also adopts self-supervision through contrastive learning~\cite{chen2020simple}.

Table~\ref{tab:models-summary} summarizes details of these models. Note that four of these models use the same network architecture -- ResNet50, yet are significantly different due to differences in their training sets, training objectives, and type of supervision.

\begin{table}[t]
\vspace{-0.1in}
\setlength\tabcolsep{4pt}
    \centering
    \scriptsize
    \begin{tabular}{l l l l } 
    
     \toprule
     \textbf{Model} & \textbf{Pretraining Dataset}& \textbf{Pretraining } & \textbf{Architecture} \\ 
     \midrule
     
     ResNet18 & ImageNet-1k (1M) & Supervised & ResNet18  \\
     
     ResNet50 & ImageNet-1k (1M) & Supervised & ResNet50\\
     
     BiT-M-R50x1 & ImageNet-21k (20M) & Supervised & ResNet50  \\ 
     
     CLIP-ViT-B/32 & 400M images from web & Supervised &Transformer\\
     MoCo ResNet50 & ImageNet-1k (1M) & Self Supervised & ResNet50\\
     SimCLR ResNet50 & ImageNet-1k (1M) & Self Supervised & ResNet50\\
     \bottomrule
    \end{tabular}
    \caption{\normalsize Summary of pretrained models tested.}
    \label{tab:models-summary}
\end{table}

\subsection{Finetuning Setup}
We evaluate the pretrained models defined in the previous section for bias before and after finetuning. We finetune the models on the COCO and on the Open Images dataset using a multi-label classification objective and a cross entropy loss. We used stochastic gradient descent with a momentum of $0.9$ and $10^{-4}$ weight decay unless indicated otherwise. We additionally finetune each model for $3$ trials with different random seeds and average results for the COCO dataset. We use as stopping criteria the mean average precision on the validation set for the COCO dataset, and the micro F1 score in the validation set for the Open Images dataset. We finetune the models until convergence. Details regarding specific hyperparameters and performance accuracy on the validation sets for each model are included in the supplementary material.

\section{Results and Discussion}
We first justify our analysis sets using image association tests and then present the results showing to what extent gender associations shift before and after fine-tuning.

\begin{figure}[t!]
    \centering
    \includegraphics[width=0.45\textwidth]{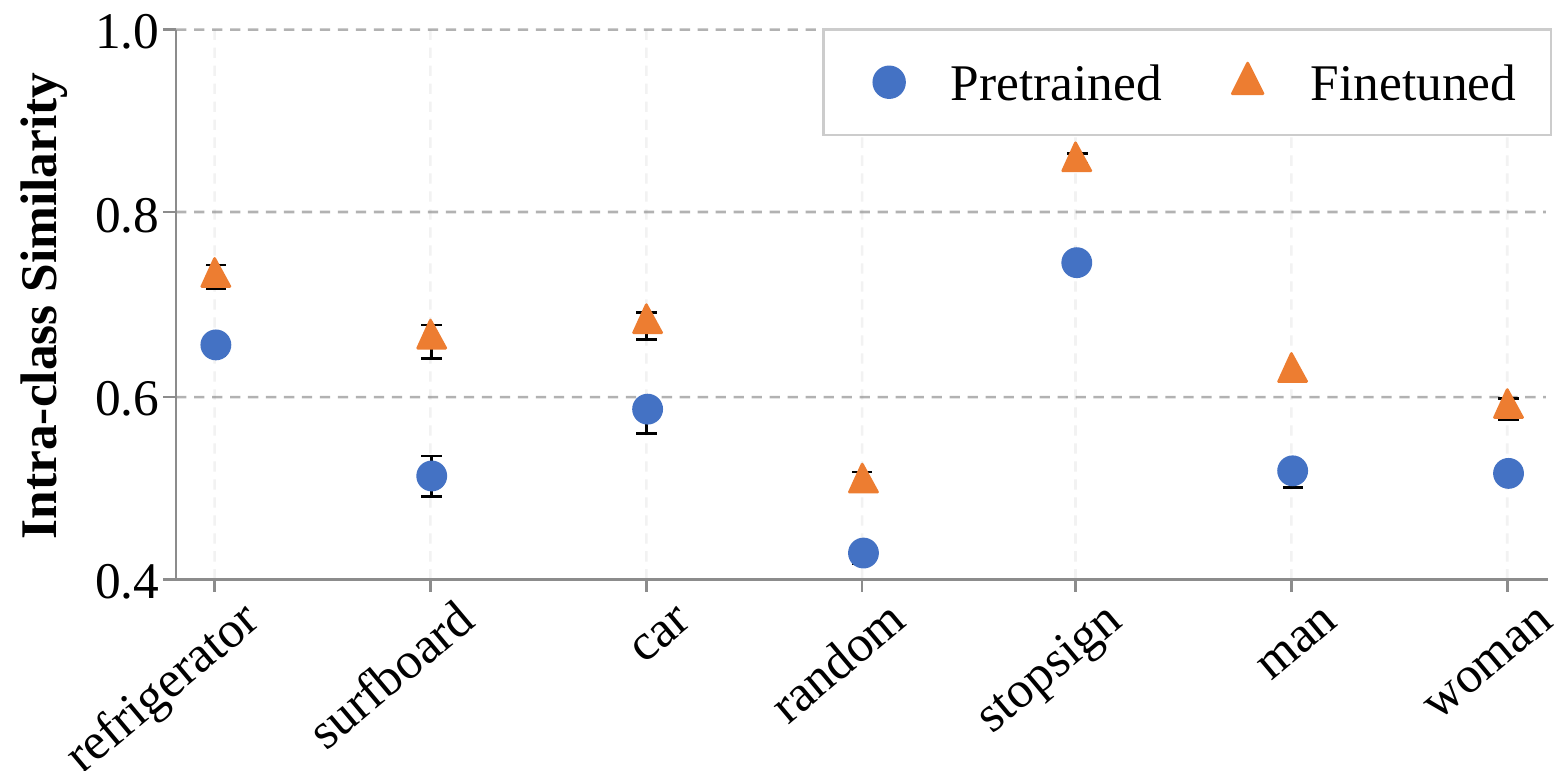} %
    \caption{Results for BiT-M-R50x1 finetuned on COCO. Intra-class similarities ($\mu_{c}$) for various individual classes where the error bars show the variation in intra-class similarities across three runs of fine-tuning. The BTS score between intra-class similarities for all {\em analysis sets} in this plot is $0.893$ with a p-value $<0.01$.}%
    \label{fig:bit_coco}%
    \vspace{-0.1in}
\end{figure}

\subsection{Image Association Tests} 
Association tests between concepts and visual estimuli have been used extensively to study human biases~\cite{greenwald2003understanding}. The recent work of Steed and Caliskan~\cite{steed2021image} adapted these tests for testing image features learned in the self-supervised setting for a variety of protected variables. In this section, we demonstrate the use of our analysis sets to perform association tests. Unlike this prior work, we measure the model's associations before and after fine-tuning for models relying on both supervised and self-supervised pretraining. Given two pairs of concepts $c_1$, e.g.~\texttt{fashion} and $c_2$, e.g.~\texttt{car}, and two attributes corresponding to protected concept categories $c_{\texttt{w}}$, e.g.~\texttt{woman} and $c_{\texttt{m}}$, e.g.~\texttt{man}. The goal is to measure whether $c_1$ is significantly closer to $c_{\texttt{w}}$ compared to the association between $c_2$ and $c_{\texttt{m}}$. In our example, we would like to test whether images in the \texttt{fashion} set hold a closer representation to images in the \texttt{woman} set on average, compared to the same relation between \texttt{car} and \texttt{man}. The goal is to measure whether the differential association between concepts $\mathbf{d}(c_{\texttt{w}}, c_{\texttt{m}}, c_1, c_2)$ as defined below is significant:
\begin{equation}
\begin{split}
    \mathbf{d} = \sum_i s(x_i^{c_1}, c_{\texttt{w}}, c_{\texttt{m}}) - \sum_j s(x_j^{c_2}, c_{\texttt{w}}, c_{\texttt{m}}),
\end{split}
\end{equation}
where $s(x, c_{\texttt{w}}, c_{\texttt{m}}) = \frac{1}{k}\sum_k\phi(x, x_k^{c_{\texttt{w}}}) - \frac{1}{t}\sum_t\phi(x, x_t^{c_{\texttt{w}}})$ measures how much closer is the representation $x$ for the current image compared to each image in the analysis sets corresponding to the protected attribute concepts. We report on Table~\ref{tab:ieat} differential scores $\mathbf{d}$ for various association tuples of the form $(c_{\texttt{w}}, c_{\texttt{m}}, c_1, c_2)$ along with $p$-values for their statistical significance under a randomized permutation test.

As observed, most models show associations that reinforce a stereotypical associations before finetuning. After fine-tuning, models that start with stereotypical associations, maintain those associations after fine-tuning, while others acquire those associations afterwards. 

Pretrained models with significantly more problematic associations before finetuning include BiT-M-R50x1 and CLIP-ViTB/32 -- we argue that this is likely due to these models being trained on a larger, noisier, and more problematic dataset~\cite{birhane2021large,birhane2021multimodal}. 
A complete set of association test results is presented in the supplementary materials.

\begin{table*}[t!] 
\setlength\tabcolsep{8pt}
\newcolumntype{Y}{>{\raggedright\arraybackslash}X}
\centering
    \small
    \begin{tabularx}{\textwidth}{l c ll c ll c ll c ll c}
     \toprule
     \multirow{2}{*}{\textbf{Model}} && \multirow{2}{*}{$c_m$} & \multirow{2}{*}{$c_w$} && \multirow{2}{*}{$c_1$} & \multirow{2}{*}{$c_2$} && \multicolumn{2}{c}{\bf Pretrained} && \multicolumn{2}{c}{\bf Finetuned}&\\
     
     \cmidrule{9-10}\cmidrule{12-13}
     
     && & &&  & && \multicolumn{1}{c}{$\mathbf{d}$} & \multicolumn{1}{c}{$p$} && \multicolumn{1}{c}{$\mathbf{d}$} & \multicolumn{1}{c}{$p$} &\\ 
     \midrule

     \multirow{2}{*}{ResNet18}  &&\texttt{man} & \texttt{woman} &&\texttt{surfboard} & \texttt{fashion} &&  0.137 & 0.114 && -0.134 & 0.874&\\
     &&\texttt{man} & \texttt{woman} &&\texttt{car} & \texttt{fashion} &&  \bandc 0.193 & \bandc 0.045 &&\bandc 0.334 &\bandc 0.002&\\
     
     \midrule

     \multirow{2}{*}{ResNet50}&&\texttt{man} & \texttt{woman} &&\texttt{surfboard} & \texttt{fashion} &&  0.0946 & 0.209 && -0.249 & 0.986&\\
     &&\texttt{man} & \texttt{woman} &&\texttt{car} & \texttt{fashion} &&  0.107 & 0.18 &&\bandc 0.372 &\bandc $10^{-4}$&\\
     
     \midrule

    \multirow{2}{*}{BiT-M-R50x1}&&\texttt{man} & \texttt{woman}  &&{\texttt{surfboard}} & \texttt{fashion} &&\bandc 0.74 &\bandc $10^{-5}$ &&\bandc 1.04 &\bandc $10^{-5}$&\\
     &&\texttt{man} & \texttt{woman} &&\texttt{car} & \texttt{fashion} &&\bandc  0.519 &\bandc $10^{-5}$ && \bandc0.996 &\bandc $10^{-5}$&\\
     
     \midrule
     
    \multirow{2}{*}{CLIP-ViT-B/32}&&\texttt{man} & \texttt{woman}  &&\texttt{surfboard} & \texttt{fashion} &&\bandc 1.031 &\bandc $10^{-5}$ &&\bandc 0.583 &\bandc $10^{-5}$&\\
     &&\texttt{man} & \texttt{woman} &&\texttt{car} & \texttt{fashion} &&\bandc 0.961 & \bandc$10^{-5}$ &&\bandc 0.763 &\bandc $10^{-5}$&\\
     
     \midrule

    \multirow{2}{*}{SimCLR ResNet50}&&\texttt{man} & \texttt{woman}  &&\texttt{surfboard} & \texttt{fashion} &&  0.146 & 0.101 && -0.267 & 0.991&\\
     &&\texttt{man} & \texttt{woman} &&\texttt{car} & \texttt{fashion} &&  0.129 & 0.126 &&\bandc 0.329 &\bandc 0.002&\\
     
     \midrule

     \multirow{2}{*}{MoCo ResNet50}&&\texttt{man} & \texttt{woman} &&\texttt{surfboard} & \texttt{fashion} &&\bandc 0.442 &\bandc $10^{-4}$ && -0.115 & 0.837&\\
     &&\texttt{man} & \texttt{woman} &&\texttt{car} & \texttt{fashion} &&\bandc 0.446 &\bandc $10^{-4}$ &&\bandc 0.427 &\bandc $10^{-4}$&\\
     \bottomrule
    \end{tabularx}
    \vspace{0.05in}
    \caption{Image association tests probing whether images in $c_{\texttt{m}}$ are closer to $c_1$ relative to $c_2$, compared to images in $c_{\texttt{w}}$. This measure is known as the differential association $\mathbf{d}$ and its statistical significance is evaluated using a randomized permutation test to obtain a $p$-value which is also reported here. The same experiments is performed for all models after fine-tuning them on a held out set in the Open Images dataset. We observe that many models already start with strong stereotypical associations even before fine-tuning them, especially those trained on larger and noisier data (e.g. CLIP).}
    \label{tab:ieat}
\end{table*}

\begin{figure}[t!]
     \centering
     \includegraphics[width=0.48\textwidth]{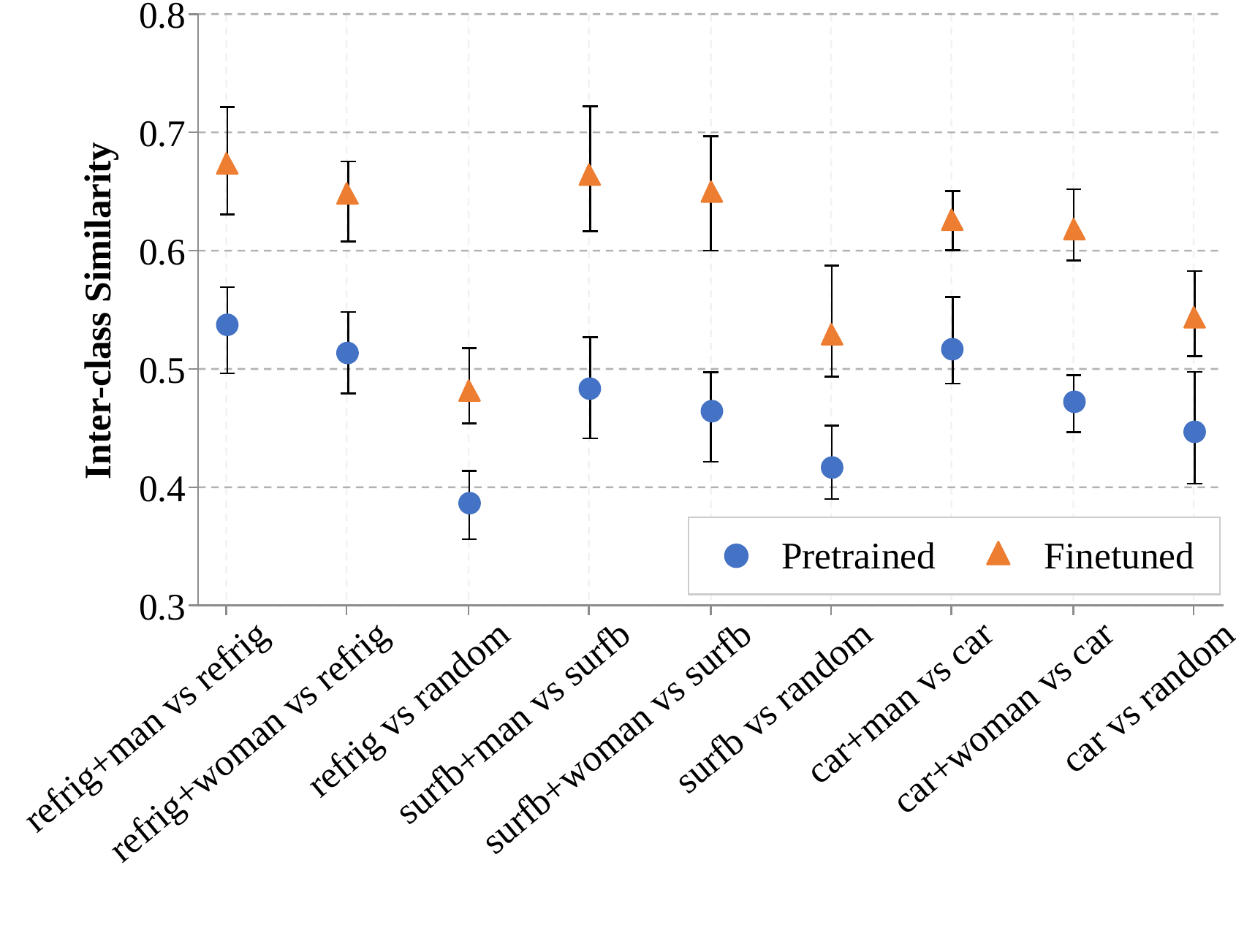} %
     \caption{Results for BiT-M-R50x1 finetuned on COCO. 
     Inter-class similarities for various pairs of {\em analysis sets} where the error bars show the variation in inter-class similarities across three runs of fine-tuning. The inter-class similarities between pairs of {\em analysis sets} in this plot have a bias transfer score of $0.791$ with a p-value $<0.01$.}%
     \label{fig:bit_coco_interclass}%
 \end{figure}

\subsection{Intra-class Similarity Transfer}
\label{sec:intra-class}
Intra-class similarity is defined in our work as the degree to which a deep neural network can cluster together images belonging to the same class in high-dimensional feature space. We quantify this intra-class similarity for class $c$ through $\mu_{c}$ as defined in Eq.~\ref{eqn:intra} using our analysis set for class $c$. 
We show in Figure~\ref{fig:bit_coco} the intra-class similarities 
for a group of COCO {\em analysis sets} computed before and after finetuning a BiT-M-R50x1 model. We include in this plot the intra-class similarity computed for a control set of random images from this dataset. As expected, the control group of random images leads to the smallest intra-class similarity score, while for a category such as stop-sign, the self-similarity is the highest both before and after fine-tuning.  It can also be observed here that even after the model has been finetuned to learn a new set of categories, a lot of the associations between concepts remain similar.  This suggests that although BiT-M-R50x1 was finetuned and its internal image representations changed, many of the associations it made before finetuning are preserved.

\subsection{Inter-class Similarity Transfer}
\label{sec:inter-class}
Inter-class similarities aim to capture the degree to which a deep neural network embeds images belonging to different classes close together in their high dimensional representation space. Figure \ref{fig:bit_coco_interclass} shows the inter-class similarities before and after fine-tuning for analysis sets obtained from COCO. Plots for all models are included in supplementary material. 
We also include controlled comparisons that compare the object class to \texttt{random} images. Here, we compute the inter-class similarity between images that contain the object and gender co-occuring with the object in isolation. As expected, the similarity between the object classes and \texttt{random} are lower than their counterparts. Similar as in with our observations in intra-class similarities BiT-M-R50x1 preserves most of the associations it already had during pretraining. \texttt{car+man vs car} has a higher inter-class similarity compared to \texttt{car+woman vs car} during pretraining but these associations have similar scores after finetuning indicating that BiT-M-R50x1 learned new image representations for \texttt{car} occurring with gender but not necessarily biased in any one direction.

\subsection{Intra-class and Inter-class Association Transfer}
In Figure~\ref{fig:bit_coco} we present intra-class similarities for BiT-M-R50x1 on COCO before and after finetuning along with its bias transfer $r_{\texttt{BTS}}$ score, and Figure~\ref{fig:bit_coco_interclass} presents inter-class similarities for the same model before and after finetuning along with its bias transfer score. Table~\ref{tab:coco_biasdown} and Table~\ref{tab:oi_biasdown} summarize the full set of bias transfer score results for all models before and after fine-tuning on both COCO and Open Images respectively. We include a full set of plots showing intra-class similarities and inter-class similarities in the supplementary material. Large scores show that models retain their biases from pretraining. As observed earlier, BiT-M-R50x1 is particularly problematic as it starts with strong gender biased associations and maintains those associations after finetuning, whereas a shallower model such as ResNet18 mostly acquires the biases of the new dataset. 
MoCo ResNet50 and ResNet18 seem to acquire a new set of biases instead of retaining the ones before finetuning. BiT-M-R50x1 and CLIP-ViT-B/32 are pretrained on large scale datasets, and their representations shift less after finetuning. More problematically BiT-M-R50x1 and CLIP-ViT-B/32 are the models with larger amounts of biases before finetuning and they are also the ones that seem to retain those biases the most.

Table~\ref{tab:oi_biasdown} shows that BiT-M-R50x1 and SimCLR ResNet50 have statistically significant BTS scores for intra-class and inter-class similarity on the Open Images {\em analysis set} retaining their pretraining biases. The intra-class BTS score for CLIP-ViT-B/32 ($0.429$) and inter-class BTS score for MoCo ResNet50 ($0.266$) show that the model adopted biases from being finetuned on the Open Images dataset where both models have adopted more biases from the Open Images dataset than from the COCO dataset as evidenced from the lower BTS scores.

\begin{table}[t!] 
\setlength\tabcolsep{5pt}
\centering
    \small
    \begin{tabularx}{0.46\textwidth}{l c ll c rr c rr c}
     \toprule
     
      && \multicolumn{2}{c}{\bf Intra-class} && \multicolumn{2}{c}{\bf Inter-class}&\\
      
     \cmidrule{3-7}\cmidrule{9-10}
     
     \multirow{2}{*}{\textbf{Model}} && \multicolumn{2}{c}{\bf $x_{i}^{c}$} && \multicolumn{2}{c}{ $x_{i}^{c_m} \text{ vs } x_{i}^{c_n}$}&\\
     
     \cmidrule{3-4}\cmidrule{6-7}\cmidrule{9-10}
     
      && \multicolumn{1}{c}{$r_{\texttt{BTS}}$} & \multicolumn{1}{c}{$p$} && \multicolumn{1}{c}{$r_{\texttt{BTS}}$} & \multicolumn{1}{c}{$p$}&\\ 
     \midrule
     
    \multirow{1}{*}{ResNet18}  && $0.714$ & $0.071$ && $0.067$ & $0.865$&\\
     

     \multirow{1}{*}{ResNet50}  && \bandc$0.857$ & \bandc$ 0.014$ && \bandc$0.867$ & \bandc$0.002$&\\
     
     \multirow{1}{*}{BiT-M-R50x1}  &&  \bandc $ 0.893$ & \bandc$0.007$ && \bandc$0.8$&  \bandc$0.01$&\\
     
     
     \multirow{1}{*}{CLIP-ViT-B/32}  && \bandc$ 0.929 $ & \bandc$0.003$ && \bandc$0.867$ & \bandc$0.002$&\\
     

     
     \multirow{1}{*}{SimCLR ResNet50}  && \bandc$0.893$ & \bandc$0.007$ && \bandc$0.817$ & \bandc$0.007$&\\
     

     \multirow{1}{*}{MoCo ResNet50}  && $0.714$ & $0.071$ && $0.367$ & $0.332$&\\
     \bottomrule
    \end{tabularx}
    \caption{
    Results using the COCO analysis sets for models finetuned on COCO. $r_{\texttt{BTS}}$ scores between the inter-class and intra-class variations across image sets. 
    }
    \label{tab:coco_biasdown}
\end{table}

\begin{table}[t!] 
\setlength\tabcolsep{5pt}
\centering
    \small
    \begin{tabularx}{0.46\textwidth}{l c ll c rr c rr c}
     \toprule
     
      && \multicolumn{2}{c}{\bf Intra-class} && \multicolumn{2}{c}{\bf Inter-class}&\\
      
     \cmidrule{3-7}\cmidrule{9-10}
     
     \multirow{2}{*}{\textbf{Model}} && \multicolumn{2}{c}{\bf $x_{i}^{c}$} && \multicolumn{2}{c}{ $x_{i}^{c_m} \text{ vs } x_{i}^{c_n}$}&\\
     
     \cmidrule{3-4}\cmidrule{6-7}\cmidrule{9-10}
     
      && \multicolumn{1}{c}{$r_{\texttt{BTS}}$} & \multicolumn{1}{c}{$p$} && \multicolumn{1}{c}{$r_{\texttt{BTS}}$} & \multicolumn{1}{c}{$p$}&\\ 
     \midrule
     \multirow{1}{*}{ResNet18}  && \bandc$ 0.81$ & \bandc$0.015$ && $0.343$ & $0.276$&\\
     

     \multirow{1}{*}{ResNet50}  && $0.667$ & $ 0.071$ && $0.545$ & $0.067$&\\
     
     \multirow{1}{*}{BiT-M-R50x1} &&  \bandc$ 0.952$ & \bandc$10^{-4}$ && \bandc$0.748$ & \bandc$0.005$&\\
     
     
     \multirow{1}{*}{CLIP-ViT-B/32}  && $  0.429 $ & $0.289$ && \bandc$0.65$ & \bandc$ 0.022$&\\
     

     
     \multirow{1}{*}{SimCLR ResNet50}  && \bandc$ 0.714 $ & \bandc$0.047$ &&  \bandc$0.671$ & \bandc$0.017$&\\
     

     \multirow{1}{*}{MoCo ResNet50} && $0.667 $ & $ 0.071$ && $0.266$ & $0.404$&\\
     \bottomrule
    \end{tabularx}
    \vspace{0.05in}
    \caption{Results using the Open Images analysis sets for models finetuned on Open Images. $r_{\texttt{BTS}}$ scores between the inter-class and intra-class variations across image sets.}
    \label{tab:oi_biasdown}
\end{table}

\section{Discussion} 
The intra-class and inter-class similarities serve as a proxy to measuring biases in the feature space for a set of classes. Tables \ref{tab:coco_biasdown} and \ref{tab:oi_biasdown} show how the models' biases are impacted after finetuning on different target datasets. The BTS score can quantify the effect of finetuning on the  intra-class and inter-class similarities and we can use this measure to determine whether a model retained its pretraining biases after finetuning.

We summarize some of our main takeaways and observations as follows:
\begin{enumerate}
    \item \textbf{Models more strongly retain their pretraining biases when finetuned on the COCO dataset in comparison to being finetuned on the Open Images dataset.} The Open Images dataset is much larger than the COCO dataset and as a result, the models are more likely to adopt the biases of the target dataset. This is a consideration to take into account when dealing with smaller target datasets. In these cases, the choice of the pretrained model is more crucial with respect to the potential biases that will need mitigation.
    \item \textbf{Finetuning on a larger dataset such as Open Images can introduce new biases as can be observed with CLIP-ViT-B/32, ResNet50 and MoCo ResNet50.} BiT-M-R50x1 and CLIP-ViT-B/32 retain their biases after finetuning on COCO but this is not necessarily the case after finetuning on Open Images. Because BiT-M-R50x1 and CLIP-ViT-B/32 are pretrained on larger datasets than COCO, they are more generalizable and thus retain their biases after finetuning but especially on a smaller dataset. 
    \item \textbf{Self-supervised training can lead to a model that has less biased associations and where less of those associations are transferred.} We observe this across our test for MoCo ResNet50 pretrained on ImageNet-1k(1M) with a self-supervised setting. However, this does not seem to be a guarantee as another self-supervised model, SimCLR ResNet50, did seem to transfer some of its associations -- however this pattern is more consistent for models trained on larger data with supervised objectives. 
\end{enumerate}

\section{Conclusion}
Our work introduced a framework for measuring the changes in the image representations computed by pretrained computer vision models before and after finetuning. We demonstrate both the effectiveness of our {\em analysis sets}, as well as the usefulness of computing correlation coefficients in the associations induced by the models on these analysis sets. We presented a detailed analysis of associations made by these models with respect to perceived gender and a set of image labels. We found that models trained with supervised objectives on larger datasets with weaker labels tend to be the ones that had the most evidence of biased associations with respect to gender. Moreover, models trained on large scale data seem to preserve their original biases more strongly after finetuning compared to models with more limited capacity trained on smaller scale data.\\\\ \textbf{Acknowledgments}
This work was supported by the NSF Program on Fairness in AI in Collaboration with Amazon under Award No IIS-2221943.

{\small
\bibliographystyle{ieee_fullname}
\bibliography{egbib}
}
\newpage 

\title{Supplementary Material} 
\author{}
\maketitle

\section{Analysis Sets}
Although it is nearly impossible to ensure the analysis sets we choose do not contain any biases, we follow existing work in curating these sets by ensuring there are no confounding scenes or objects in the images for a given category. We also ensure that the images have similar backgrounds to limit variability and unforeseen biases in the analysis sets. We primarily want to examine gender bias and thus we choose categories that allow us to compare the co-occurence of gender annotations with different objects. For example, if we want to examine the biases of categories \texttt{man} and \texttt{woman} with respect to an object such as a \texttt{car}, our analysis set would need to include at least five classes: [\texttt{man}, \texttt{woman}, \texttt{car}, \texttt{man+car}, \texttt{woman+car}, \texttt{random}] where the gender+object categories include images containing the people with a perceived gender and the object, and ideally no other objects. We also include a category labeled as \texttt{random} that has a random subset of images from the dataset and serves as a reference for comparison with our categories of interest. 
\subsection{COCO Analysis Set}
\begin{figure*}[t!]
    \centering {{\includegraphics[width=\textwidth]{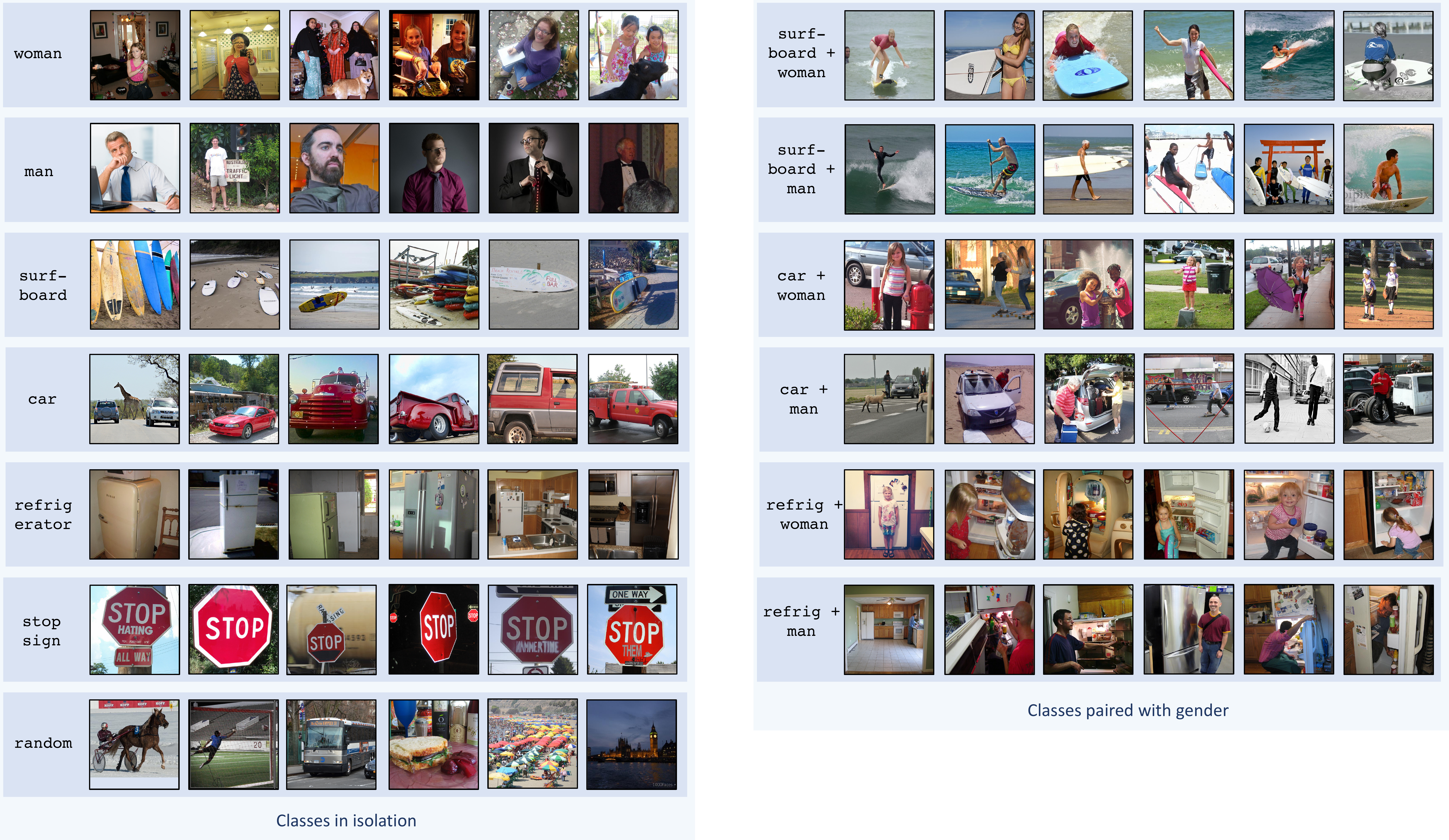} }}%
    \caption{Quantitative examples from COCO \emph{analysis set}}
    \label{fig:coco-analysis}%
\end{figure*}

\begin{wraptable}{r}{3cm}
\setlength\tabcolsep{5pt}
    \centering
    \scriptsize
    \begin{tabular}{l r} 
    
     \toprule
     \textbf{Class}  & \textbf{n} \\ [0.5ex] 
     \midrule
     Man & $12$  \\ 
     Woman  & $15$ \\
     Random & $20$\\
     Stop Sign & $44$ \\
     Car & $12$ \\
     Car+man & $9$ \\
     Car+woman & $6$ \\
     Refrigerator & $18$ \\
     Refrigerator+man & $8$ \\
     Refrigerator+woman & $9$ \\
     Surfboard & $14$ \\
     Surfboard+man & $23$ \\
     Surfboard+woman & $16$ \\
     \bottomrule
    \end{tabular}
    \vspace{0.05in}
    \caption{COCO Analysis Set statistics.}
    \label{tab:coco_analysis}
\end{wraptable}
Table \ref{tab:coco_analysis} provides an overview of the classes and the number of examples $n$ in each class for the COCO \emph{analysis set}. The COCO \emph{analysis set} was collected using a combination of object annotations and manual inspection to ensure each image is a representative sample for each individual concept. Figure~\ref{fig:coco-analysis} shows sample images for classes that depict a single concept in isolation and for classes that show a concept co-occurring with a gender.

\subsection{Open Images Analysis Set}
\begin{figure*}[t!]
    \centering {{\includegraphics[width=\textwidth]{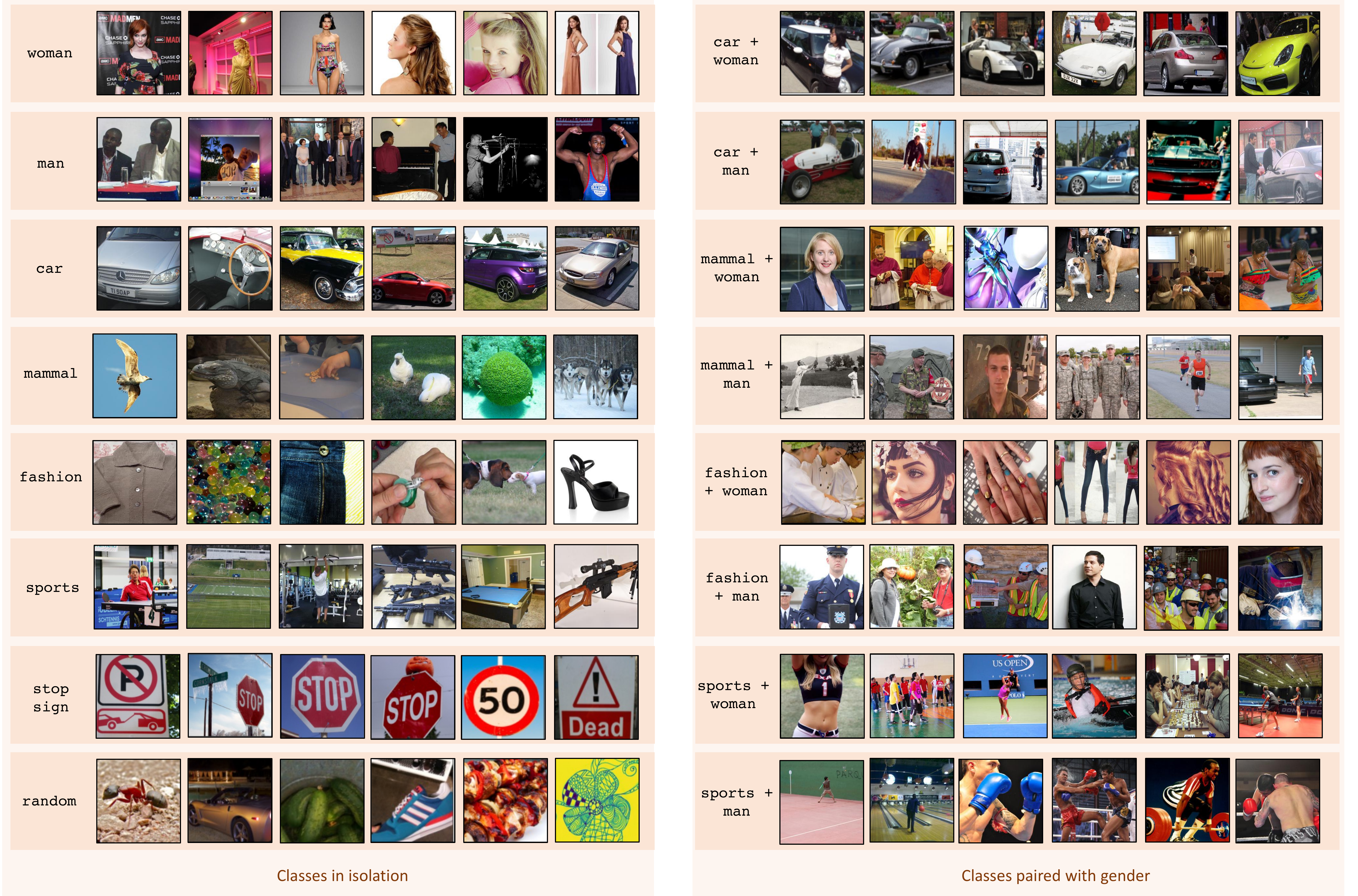} }}%
    \caption{Quantitative examples from Open Images \emph{analysis set}}
    \label{fig:oi-analysis}%
\end{figure*}

\begin{wraptable}{r}{3cm}
\setlength\tabcolsep{4pt}
    \centering
    \scriptsize
    \vspace{-0.5in}
    \begin{tabular}{l r}  
    
     \toprule
     \textbf{Class}  & \textbf{n} \\ [0.5ex] 
     \midrule
     Man & $150$  \\ 
     Woman  & $150$ \\
     Random & $150$\\
     Stop Sign & $22$ \\
     Car & $150$ \\
     Car+man & $150$ \\
     Car+woman & $49$ \\
     Sports & $150$ \\
     Sports+man & $120$ \\
     Sports+woman & $32$ \\
     Fashion & $150$ \\
     Fashion+man & $52$ \\
     Fsahion+woman & $150$ \\
     Mammal & $150$ \\
     Mammal+man & $150$ \\
     Mammal+woman & $150$ \\
     \bottomrule
    \end{tabular}
    \vspace{0.05in}
    \caption{Openimages v4 Analysis Set statistics.}
    \label{tab:oi_analysis}
\end{wraptable}
Table \ref{tab:oi_analysis} provides an overview of the classes and the number of examples $n$ in each class in the Open Images \emph{analysis set}. The Open Images \emph{analysis set} was also in a more automated way than the COCO \emph{analysis set} by only using the object annotations for each image. As a result, each class in the Open Images \emph{analysis set} has a lot more examples than the COCO \emph{analysis set}. Figure~\ref{fig:oi-analysis} shows examples for images from the Open Images analysis sets. 

\vspace{-0.1in}
\section{Experimental Details}

We finetune each model on the COCO 2017 and Open Images datasets on a multi-label object classification task. Table \ref{tab:coco_ft} provides details on the mean average precision score for each model finetuned on COCO. Unlike the other models, ResNet18 and MoCo ResNet50 are finetuned with the \emph{ReduceLROnPlateau} learning rate scheduler and weight decay and SimCLR ResNet50 is trained with the LARS scheduler. Each model is finetuned on COCO for at least $3$ trials. Table \ref{tab:oi_finetune} provides details on the micro F1 score for each model finetuned on Open Images. Each model is finetuned for a single trial on Open Images to a comparable F1 score. MoCo ResNet50 is trained with the \emph{ReduceLROnPlateau} learning rate scheduler and SimCLR ResNet50 is trained with the LARS scheduler. All models are trained with SGD optimizer with a momentum of 0.9 and weight decay except for SimCLR ResNet50 which is trained without weight decay. 

\begin{table}[bh]
\setlength\tabcolsep{4pt}
    \centering
    \scriptsize
    \begin{tabular}{l c l l c} 
    
     \toprule
     \textbf{Model}  & \textbf{Epochs} & \textbf{Learning Rate} & \textbf{Optimizer} & \textbf{mAP} \\ [0.5ex] 
     \midrule
      BiT-M-R50x1 & 25 & 0.003 & SGD, m: 0.9 & 0.744 $\pm$ 0.001  \\ 
     ResNet50  & 25 & 0.001  & SGD, m: 0.9 & 0.635 $\pm$  0.002\\
     ResNet18 & 25 & 0.1 & SGD, m: 0.9, & 0.643 $\pm$  0.003  \\
     CLIP: ViT-B/32 & 25 & 0.001  & SGD, m: 0.9 & 	0.705 $\pm$ 0.009\\
     SimCLR ResNet50 & 25 & 0.01  & SGD, m: 0.9 & 0.573 $\pm$ 0.001\\
     MoCo ResNet50 & 25 & 0.1  & SGD, m: 0.9, & 0.632 $\pm$ 0.038\\
     \bottomrule
    \end{tabular}
    \vspace{0.05in}
    \caption{Results and details for  finetuning models on COCO 2017.}
    \label{tab:coco_ft}
\end{table}
\begin{table}[bh]
\setlength\tabcolsep{4pt}
    \centering
    \scriptsize
    \begin{tabular}{l c l l c} 
    \toprule 
     \textbf{Model}  & \textbf{Epochs} & \textbf{Learning Rate} & \textbf{Optimizer} & \textbf{Micro F1} \\  [0.5ex] 
     \midrule
      BiT-M-R50x1 & 15 & 0.001 & SGD, m: 0.9 & 0.417\\
     ResNet50 & 15 & 0.001 & SGD, m: 0.9 & 0.329\\
     ResNet18 & 30 & 0.001  & SGD, m: 0.9 & 0.311 \\
     CLIP: ViT-B/32 & 15 & 0.001  & SGD, m: 0.9& 0.396\\
     SimCLR ResNet50 & 15 & 0.01  & SGD, m: 0.9 & 0.284\\
     MoCo ResNet50 & 15 & 0.1  & SGD, m: 0.9 & 0.308\\ 
     \bottomrule
    \end{tabular}
    \vspace{0.05in}
    \caption{Results and details for finetuning models on Open Images v4.} 
    \label{tab:oi_finetune}
\end{table}
\section{iEAT -- Full Experiment}

Table \ref{tab:ieat} provides a complete set of results from replicating the iEAT experiment on the Open Images \emph{analysis set}. We continue to observe that SimCLR ResNet50 and MoCo ResNet50 do not show any statistically significant effect sizes for any concept pairings. Additionally, we observe that BiT-M-R50x1 starts with a moderate stereotypical association ($d=0.61$ for \texttt{\{mammal, fashion\}}) and that association is reduced after finetuning but still remains moderate. Similarly, ResNet18 and ResNet50 also develop a low to moderate association along this same relation after finetuning. On the contrary, CLIP-ViT-B/32 starts with a stronger stereotypical association for \texttt{\{mammal, fashion\}} and this is not mirrored after finetuning. Both BiT-M-R50x1 and CLIP-ViT-B/32 start with low stereotypical associations for \texttt{\{surfboard, car\}} at ($d=0.35$ and $d=0.27$, respectively) but this association is not maintained after finetuning. For the \texttt{\{car, mammal\}} relation, we observe that both BiT-M-R50x1 and CLIP-ViT-B/32 develop strong stereotypical associations after finetuning ($d=0.79$ and $d=0.69$, respectively). For BiT-M-R50x1, CLIP-ViT-B/32 and ResNet50, we also observe that all these models develop some level of association after finetuning for \texttt{\{car, mammal\}}. 
\begin{table*}[th!] 
\setlength\tabcolsep{14pt}
\newcolumntype{Y}{>{\raggedright\arraybackslash}X}
\centering
    \scriptsize
    \begin{tabularx}{\textwidth}{l c ll c ll c ll c}
     \toprule
     \multirow{2}{*}{\textbf{Model}} && \multirow{2}{*}{\textbf{Concept-A}} & \multirow{2}{*}{\textbf{Concept-B}} && \multicolumn{2}{c}{\bf Pretrained} && \multicolumn{2}{c}{\bf Finetuned}&\\
     
     \cmidrule{6-7}\cmidrule{9-10}
     
      &&  & && \multicolumn{1}{c}{\textit{Effect Sizes}} & \multicolumn{1}{c}{\textit{p}} && \multicolumn{1}{c}{\textit{Effect Sizes}} & \multicolumn{1}{c}{\textit{p}} &\\ 
     \midrule
     \multirow{6}{*}{BiT-M-R50x1} && Surfboard & Car &&  \bandc 0.346 & \bandc $10^{-4}$ && -0.061 & 0.702&\\
     &&{Surfboard} & Fashion &&\bandc 0.74 &\bandc $10^{-5}$ &&\bandc 1.04 &\bandc $10^{-5}$&\\
     &&Surfboard & Mammal &&\bandc 0.519 &\bandc $10^{-5}$ &&\bandc 0.809 &\bandc $10^{-5}$&\\
     &&Car & Fashion &&\bandc  0.519 &\bandc $10^{-5}$ && \bandc0.996 &\bandc $10^{-5}$&\\
     &&Car & Mammal &&  0.091 & 0.219 &&\bandc 0.788 &\bandc $10^{-5}$&\\
     &&Mammal & Fashion &&\bandc  0.605 &\bandc $10^{-5}$ &&\bandc 0.508 &\bandc $10^{-5}$&\\
     
     \midrule
     
     \multirow{6}{*}{CLIP-ViT-B/32} && Surfboard & Car && \bandc 0.272  & \bandc0.01 && -0.447 & 0.999&\\
     &&Surfboard & Fashion &&\bandc 1.031 &\bandc $10^{-5}$ &&\bandc 0.583 &\bandc $10^{-5}$&\\
     &&Surfboard & Mammal &&\bandc 0.558 &\bandc $10^{-5}$ &&\bandc 0.551 &\bandc $10^{-5}$&\\
     &&Car & Fashion &&\bandc 0.961 & \bandc$10^{-5}$ &&\bandc 0.763 &\bandc $10^{-5}$&\\
     &&Car & Mammal && \bandc 0.298 & \bandc 0.004 &&\bandc 0.687 &\bandc $10^{-5}$&\\
     &&Mammal & Fashion &&\bandc 0.84 &\bandc $10^{-5}$ && -0.249 & 0.984&\\
     
     \midrule
     
     \multirow{6}{*}{ResNet18} && Surfboard & Car && -0.169 & 0.932 && -0.303 & 0.995&\\
     &&Surfboard & Fashion &&  0.137 & 0.114 && -0.134 & 0.874&\\
     &&Surfboard & Mammal && 0.101 & 0.192 && -0.332 & 0.998&\\
     &&Car & Fashion &&  \bandc 0.193 & \bandc 0.045 &&\bandc 0.334 &\bandc 0.002&\\
     &&Car & Mammal &&  0.165 & 0.076 && 0.188 & 0.054&\\
     &&Mammal & Fashion &&  $-10^{-4}$ & 0.503 && \bandc 0.199 & \bandc 0.041&\\
     
     \midrule

     \multirow{6}{*}{ResNet50} && Surfboard & Car && -0.084 & 0.77 && -0.341 & 0.999&\\
     &&Surfboard & Fashion &&  0.0946 & 0.209 && -0.249 & 0.986&\\
     &&Surfboard & Mammal &&  0.041 & 0.354 && -0.278 & 0.992&\\
     &&Car & Fashion &&  0.107 & 0.18 &&\bandc 0.372 &\bandc $10^{-4}$&\\
     &&Car & Mammal && 0.068 & 0.272 && \bandc 0.265 & \bandc 0.01&\\
     &&Mammal & Fashion &&  0.022 & 0.43 && \bandc 0.219 & \bandc 0.03&\\
     
     \midrule
     
     \multirow{6}{*}{SimCLR ResNet50} && Surfboard & Car && -0.029 & 0.604 && -0.32 & 0.998&\\
     &&Surfboard & Fashion &&  0.146 & 0.101 && -0.267 & 0.991&\\
     &&Surfboard & Mammal &&  0.139 & 0.115 && -0.08 & 0.748&\\
     &&Car & Fashion &&  0.129 & 0.126 &&\bandc 0.329 &\bandc 0.002&\\
     &&Car & Mammal &&  0.097 & 0.189 &&\bandc 0.281 &\bandc 0.007&\\
     &&Mammal & Fashion &&  0.072 & 0.275 && -0.14 & 0.884&\\
     
     \midrule

     \multirow{6}{*}{MoCo ResNet50} && Surfboard & Car && -0.239 & 0.983 && -0.456 & 1.0&\\
     &&Surfboard & Fashion &&\bandc  0.442 &\bandc $10^{-4}$ && -0.115 & 0.837&\\
     &&Surfboard & Mammal &&\bandc 0.247 &\bandc 0.018 && -0.032 & 0.609&\\
     &&Car & Fashion &&\bandc 0.446 &\bandc $10^{-4}$ &&\bandc 0.427 &\bandc $10^{-4}$&\\
     &&Car & Mammal && \bandc 0.298 &\bandc 0.005 &&\bandc 0.429 &\bandc $10^{-4}$&\\
     &&Mammal & Fashion &&\bandc 0.243 &\bandc 0.02 && -0.106 & 0.82&\\ 
     \bottomrule
    \end{tabularx}
    \vspace{0.05in}
    \caption{iEAT Experiment on Open Images Analysis Set.  We replicate the iEAT test for associations the model makes between target concepts: man and woman, and attributes from the Open Images Analysis Set such as surfboard, mammal, etc. We test these associations on the features generated from pretrained and finetuned models.  Effect sizes represent the magnitude of bias, and the p-values show the significance of the permutation test. }
    \vspace{0.1in}
    \label{tab:ieat}
\end{table*}

\section{COCO Analysis Set - Full Experiments}
\begin{figure*}[h]
    \centering {{\includegraphics[width=0.85\textwidth]{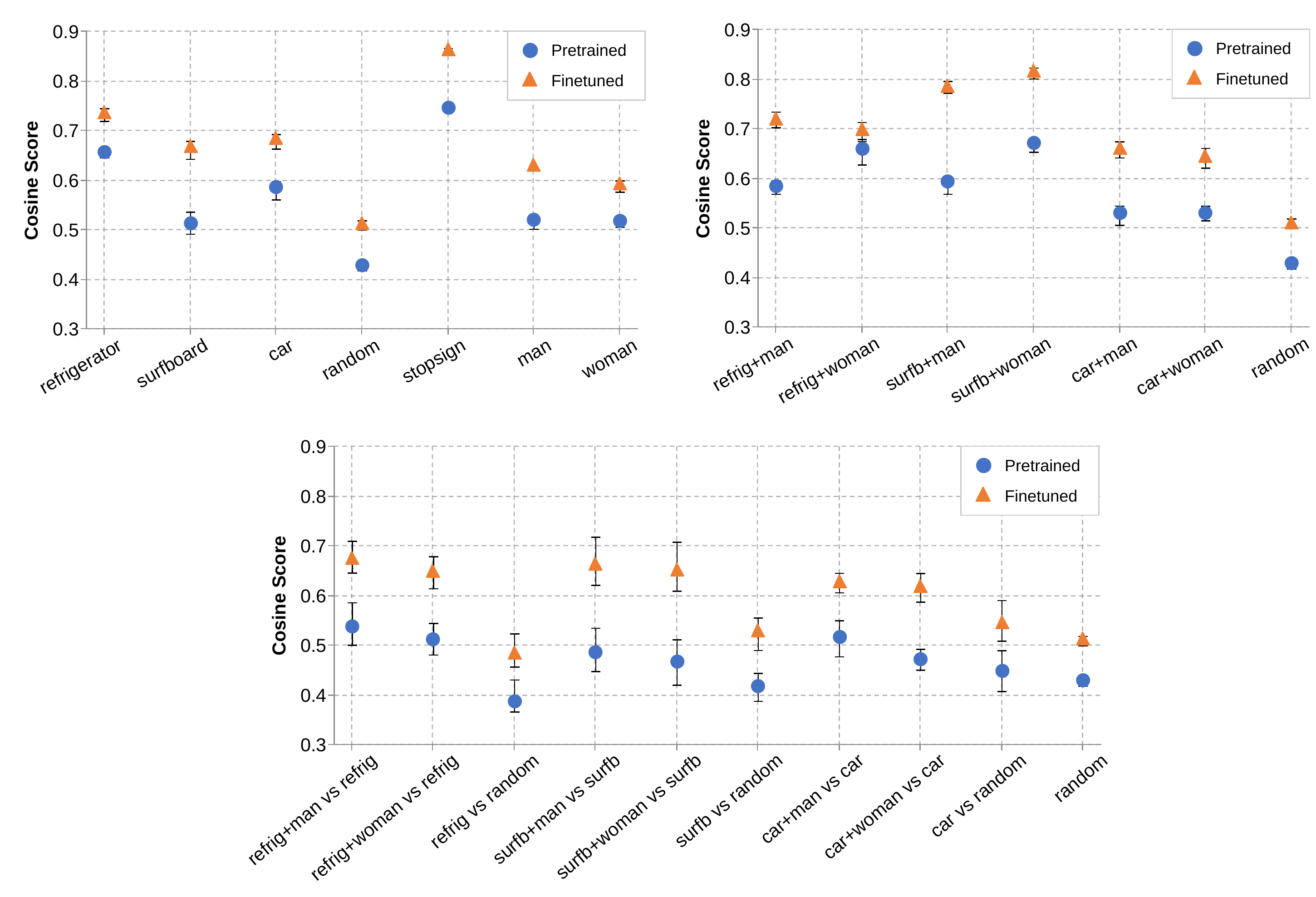} }}%
    \caption{BiT-M-R50x1 finetuned on COCO and evaluated on COCO analysis set.}
    \label{fig:coco_bit}%
\end{figure*}

\begin{figure*}[h]
    \centering {{\includegraphics[width=0.85\textwidth]{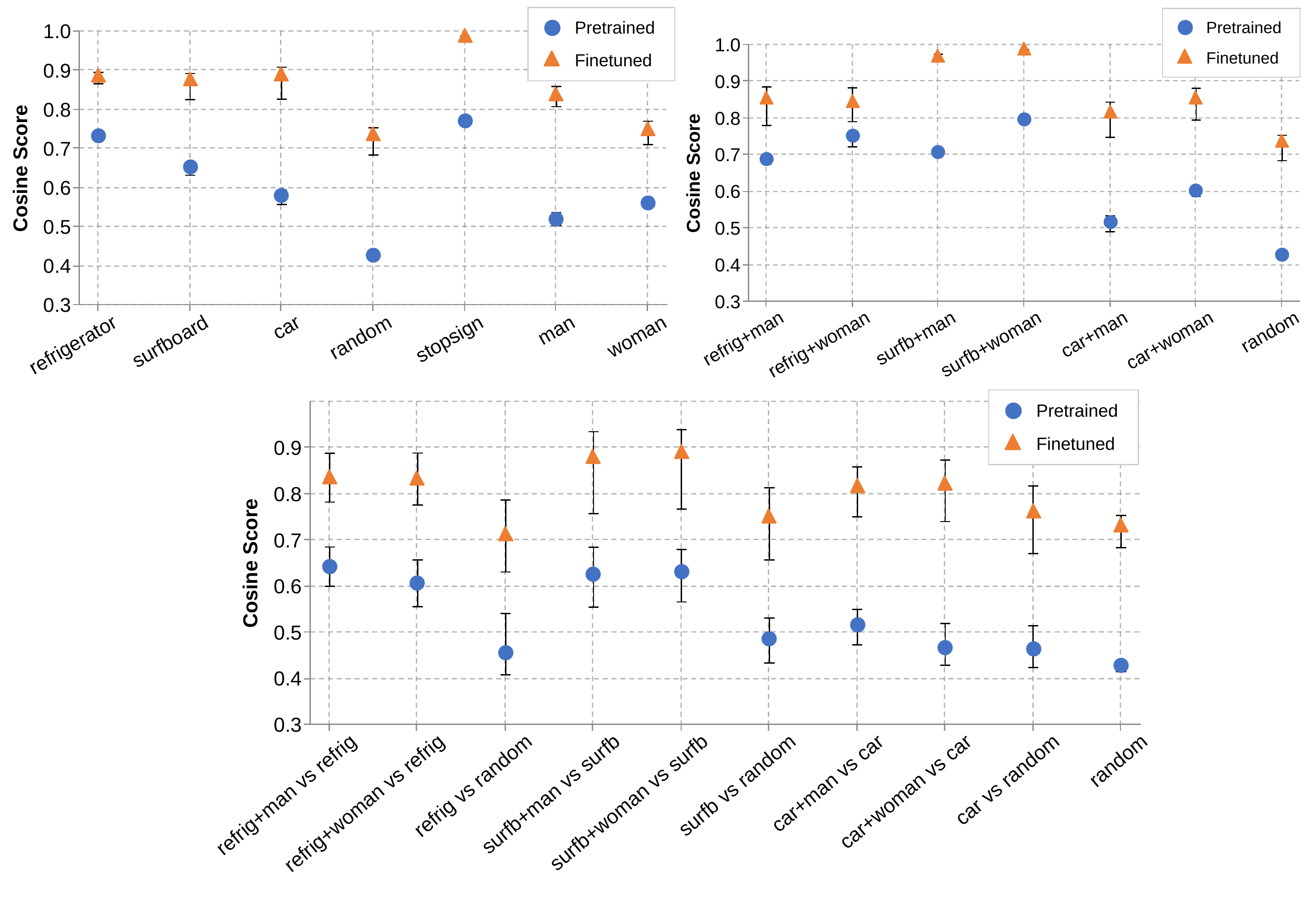} }}%
    \caption{CLIP-ViT-B/32 finetuned on COCO and evaluated on COCO analysis set.}
    \label{fig:coco_clip}%
\end{figure*}

\begin{figure*}[h]
    \centering {{\includegraphics[width=0.85\textwidth]{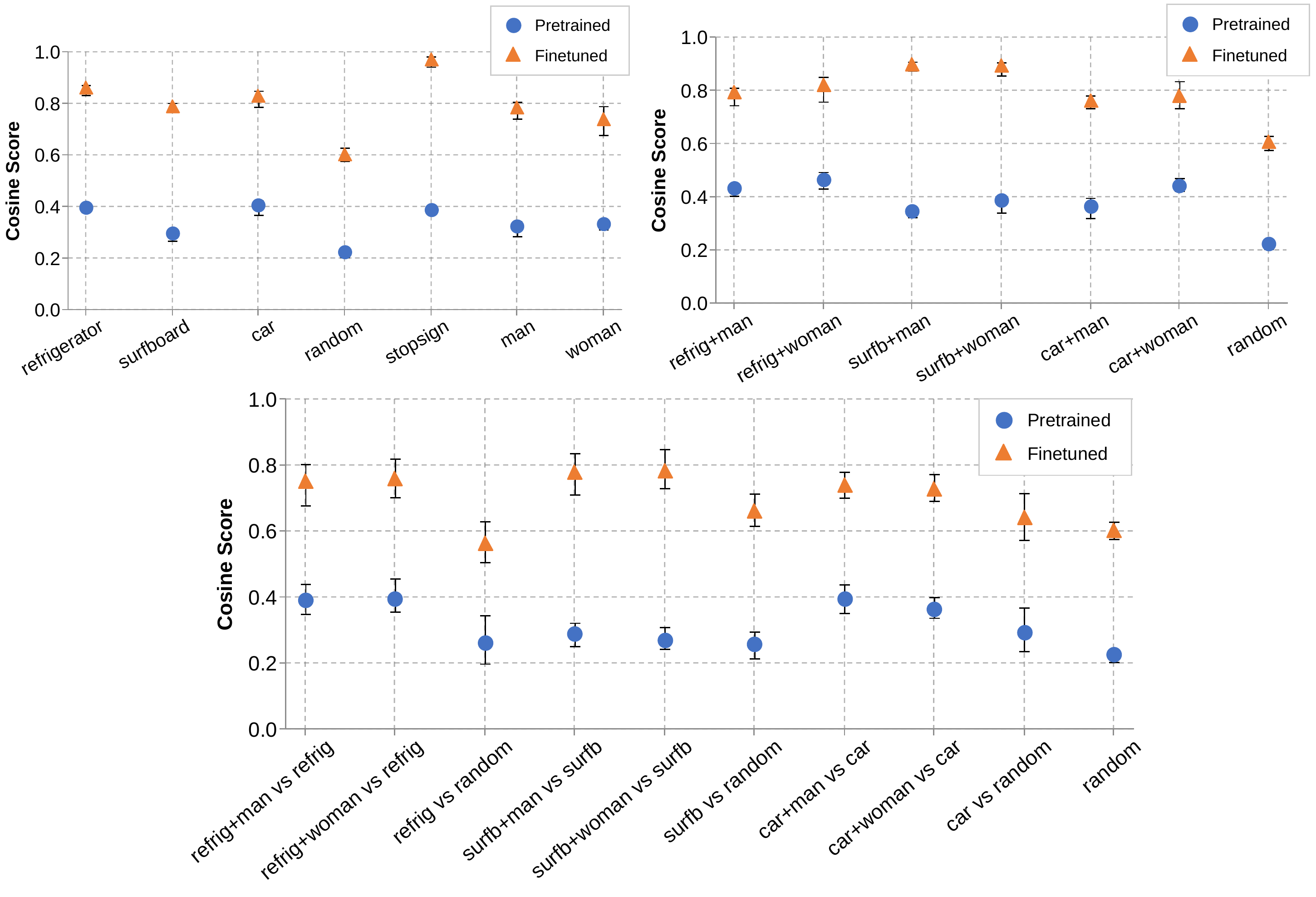} }}%
    \caption{MoCo ResNet50 finetuned on COCO and evaluated on COCO analysis set.}
    \label{fig:coco_moco}%
\end{figure*}

\begin{figure*}[h]
    \centering {{\includegraphics[width=0.85\textwidth]{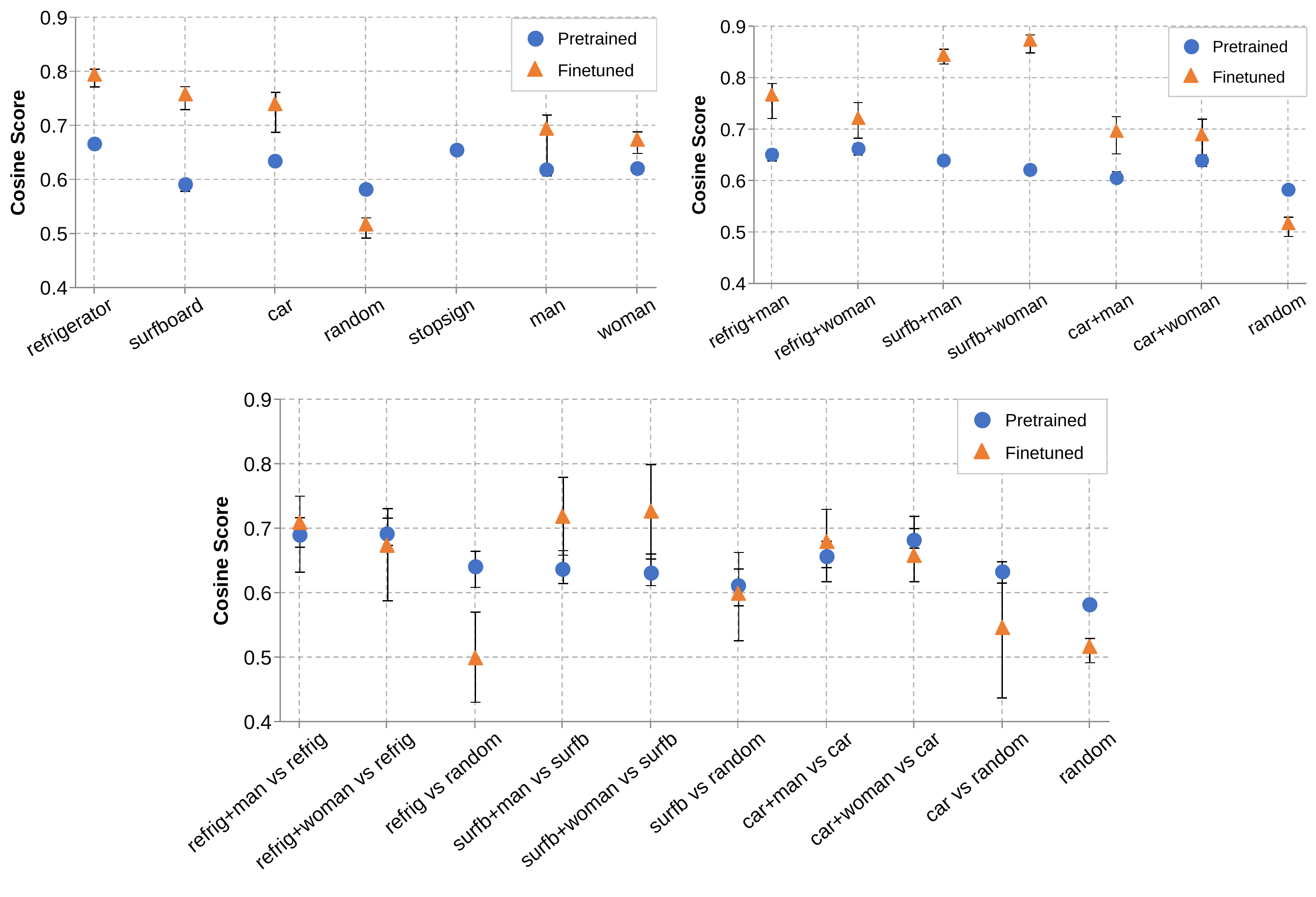} }}%
    \caption{ResNet18 finetuned on COCO and evaluated on COCO analysis set.}
    \label{fig:coco_resnet18}%
\end{figure*}

\begin{figure*}[h]
    \centering {{\includegraphics[width=0.85\textwidth]{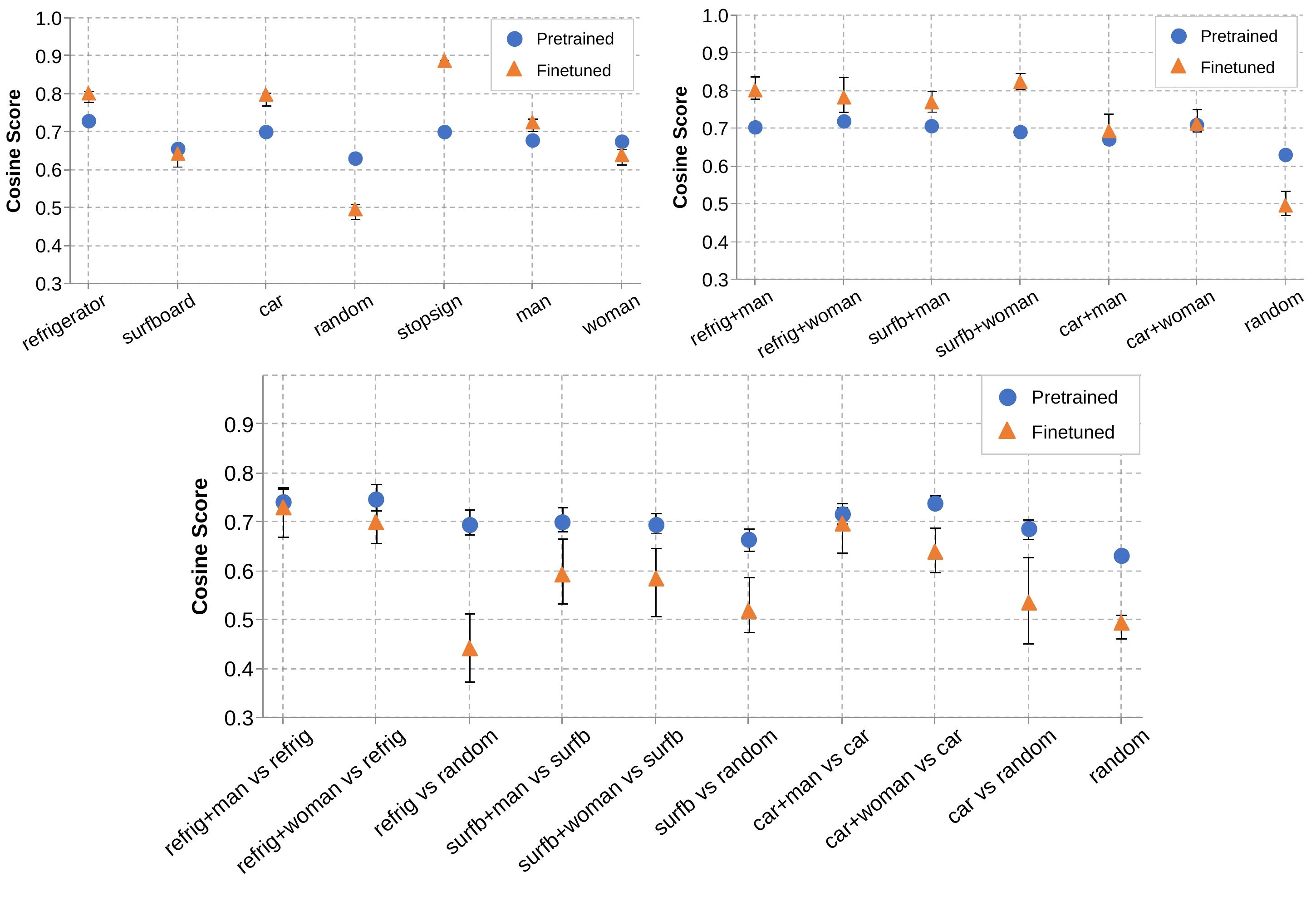} }}%
    \caption{ResNet50 finetuned on COCO and evaluated on COCO analysis set.}
    \label{fig:coco_resnet50}%
\end{figure*}

\begin{figure*}[h]
    \centering {{\includegraphics[width=0.85\textwidth]{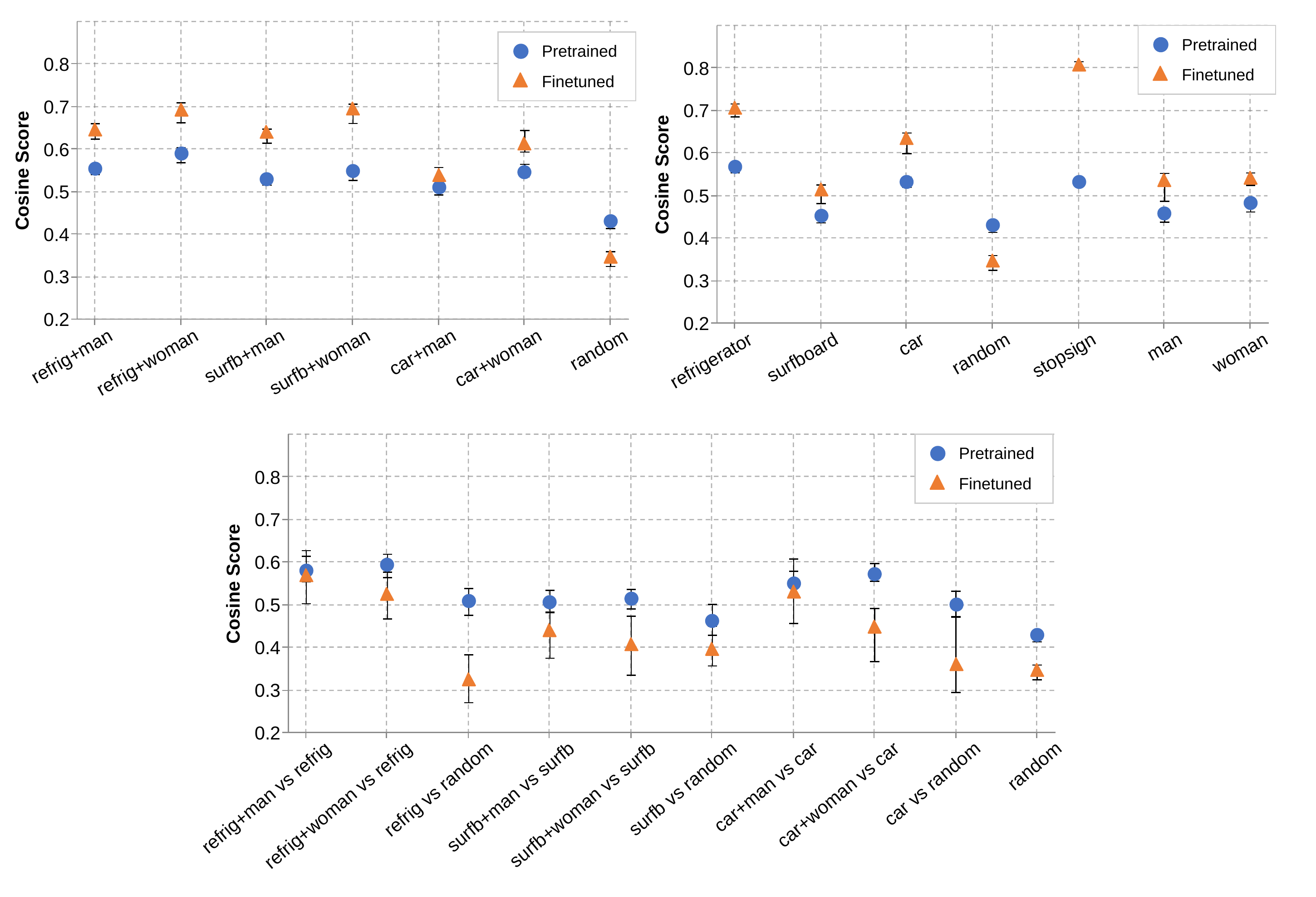} }}%
    \caption{SimCLR ResNet50 finetuned on COCO and evaluated on COCO analysis set.}
    \label{fig:coco_simclr}%
\end{figure*}

We provide a complete set of results for each model finetuned on COCO 2017 and evaluated on the COCO \emph{analysis set}. Figures $ [ \ref{fig:coco_bit}, \ref{fig:coco_clip}, \ref{fig:coco_moco}, \ref{fig:coco_resnet18}, \ref{fig:coco_resnet50},\ref{fig:coco_simclr} ]$ show the intra-class (top row) and inter-class similarities (bottom row) before and after finetuning a model on COCO. We observe in Figures [\ref{fig:coco_bit}, \ref{fig:coco_clip}, \ref{fig:coco_resnet50}, \ref{fig:coco_simclr}] that BiT-M-R50x1, CLIP-ViT-B/32, ResNet50 and SimCLR ResNet50 preserve a lot of the associations even after finetuning as can be noted by the similar trends between pretraining and finetuning similarities. We also note that ResNet18 and MoCo ResNet50 acquire new biases after finetuning as can be observed from lower correlations between in Figures \ref{fig:coco_moco} and \ref{fig:coco_resnet18}. For example, Figure \ref{fig:coco_resnet18} shows that ResNet18 has a intra-class higher similarity score for \texttt{surfboard + man} than it does for \texttt{surfboard + woman} before finetuning but these scores flip after finetuning showing that ResNet18 acquired new biases during the finetuning stage. ResNet18 and MoCo ResNet50 are both trained on smaller scale datasets than CLIP-ViT-B/32 and BiT-M-R50x1 and as a result, their biases are more prone to shifting after finetuning. 

\section{Open Images Analysis Set - Full Experiments}
\begin{figure*}[h]
    \centering {{\includegraphics[width=0.85\textwidth]{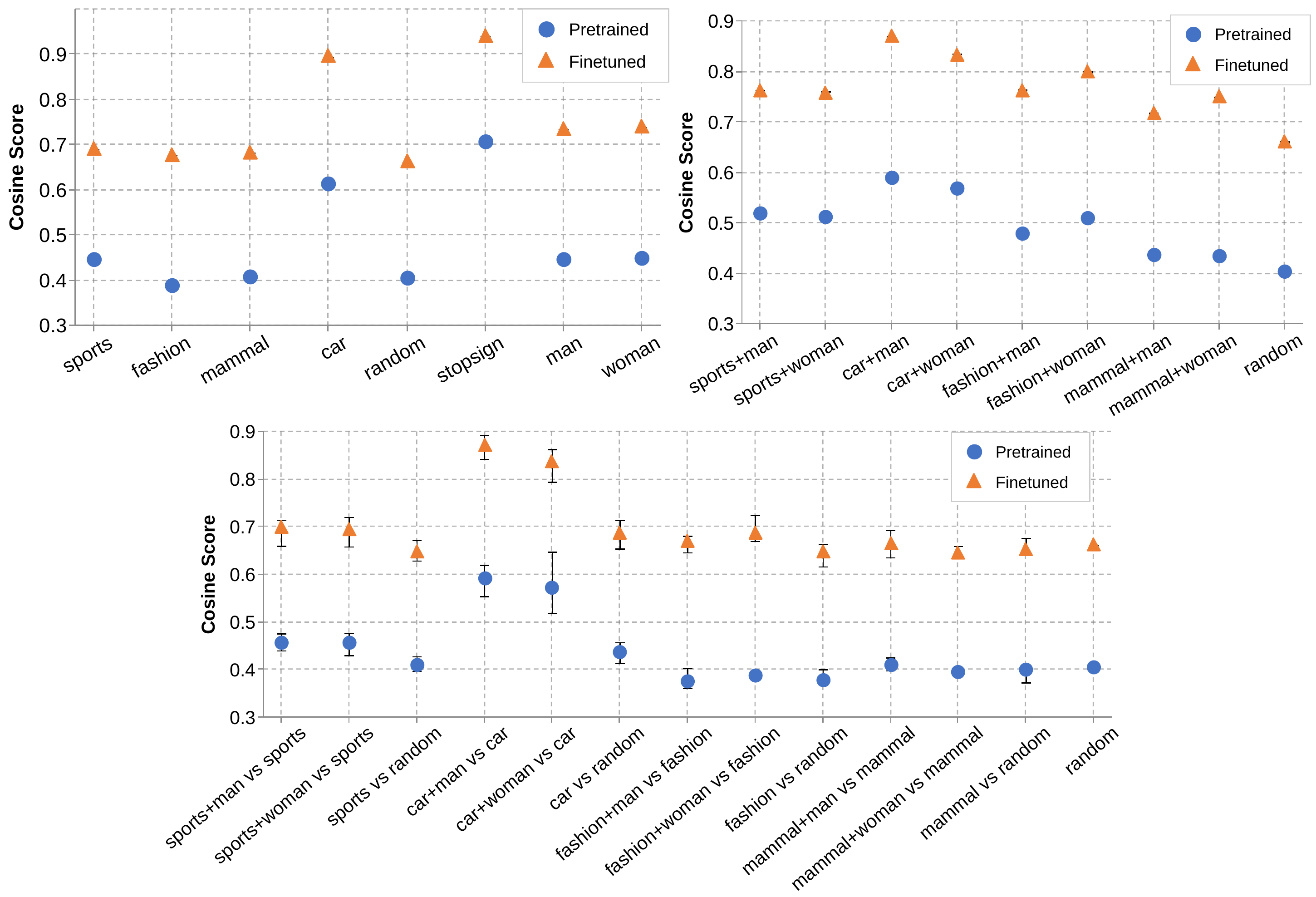} }}%
    \caption{BiT-M-R50x1 finetuned on Open Images and evaluated on Open Images analysis set.}
    \label{fig:openimages_bit}%
\end{figure*}

\begin{figure*}[h]
    \centering {{\includegraphics[width=0.85\textwidth]{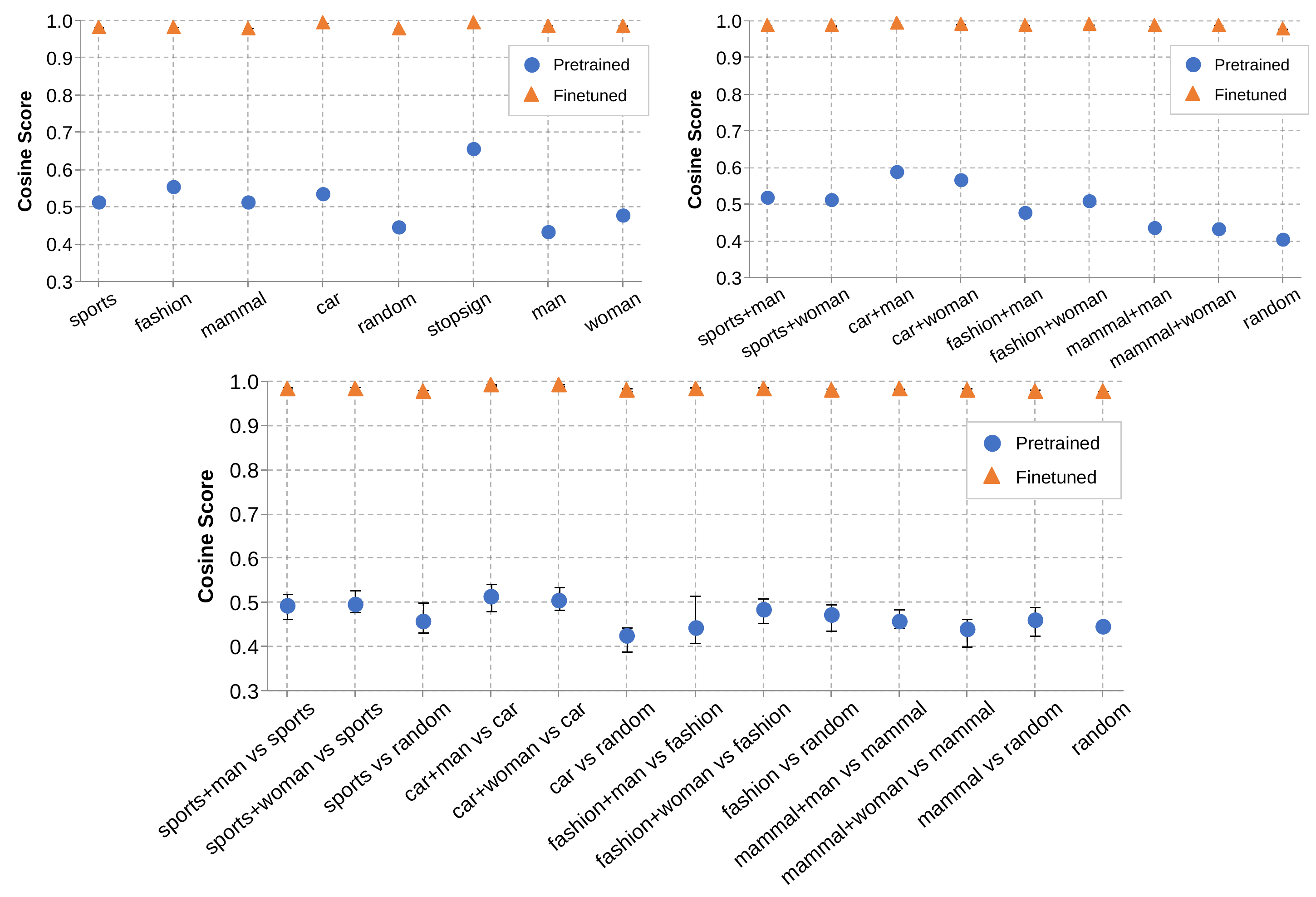} }}%
    \caption{CLIP-ViT-B/32 finetuned on Open Images and evaluated on Open Images analysis set.}
    \label{fig:openimages_clip}%
\end{figure*}

\begin{figure*}[h]
    \centering {{\includegraphics[width=0.85\textwidth]{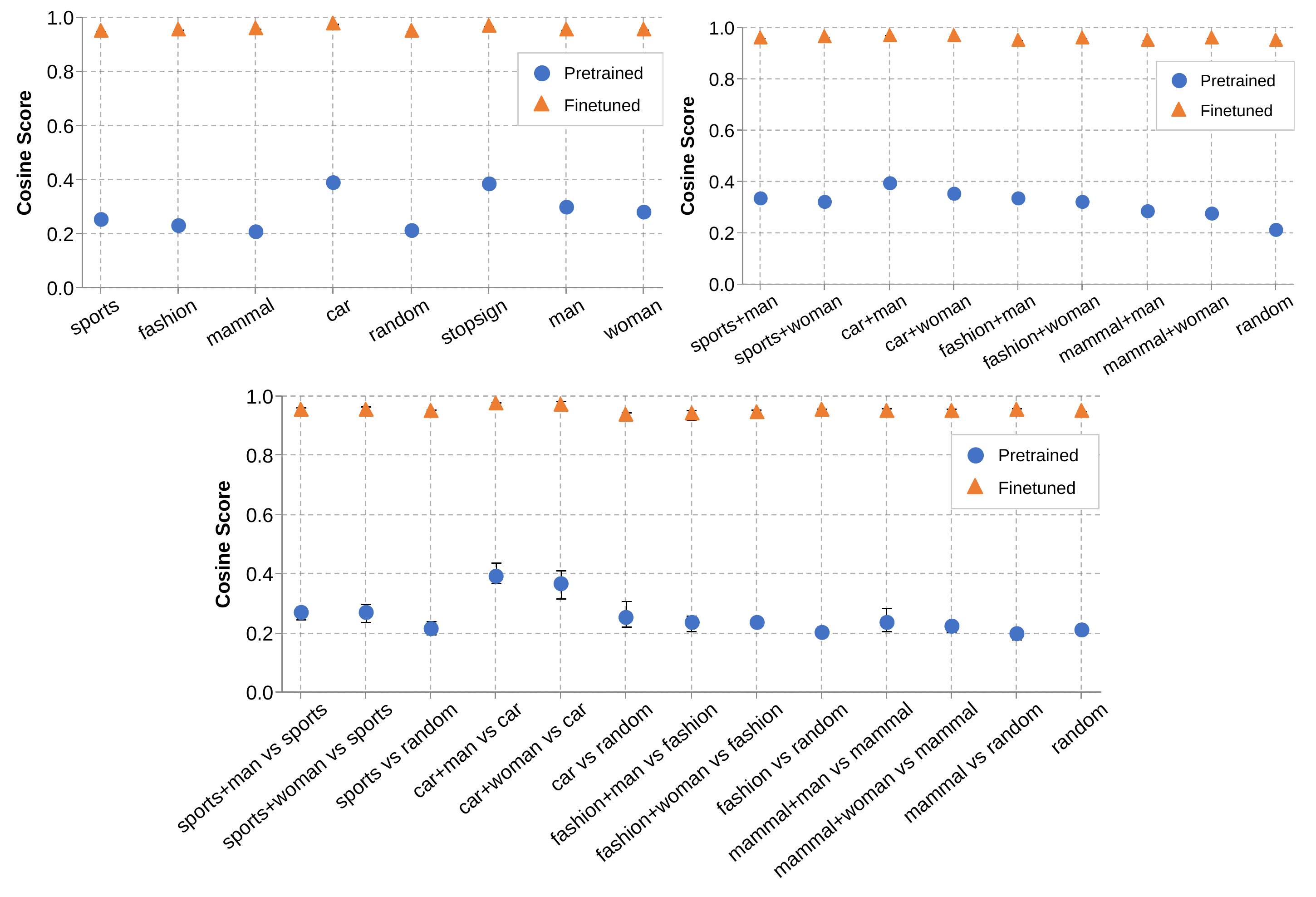} }}%
    \caption{MoCo ResNet50 finetuned on Open Images and evaluated on Open Images analysis set.}
    \label{fig:openimages_moco}%
\end{figure*}

\begin{figure*}[h]
    \centering {{\includegraphics[width=0.85\textwidth]{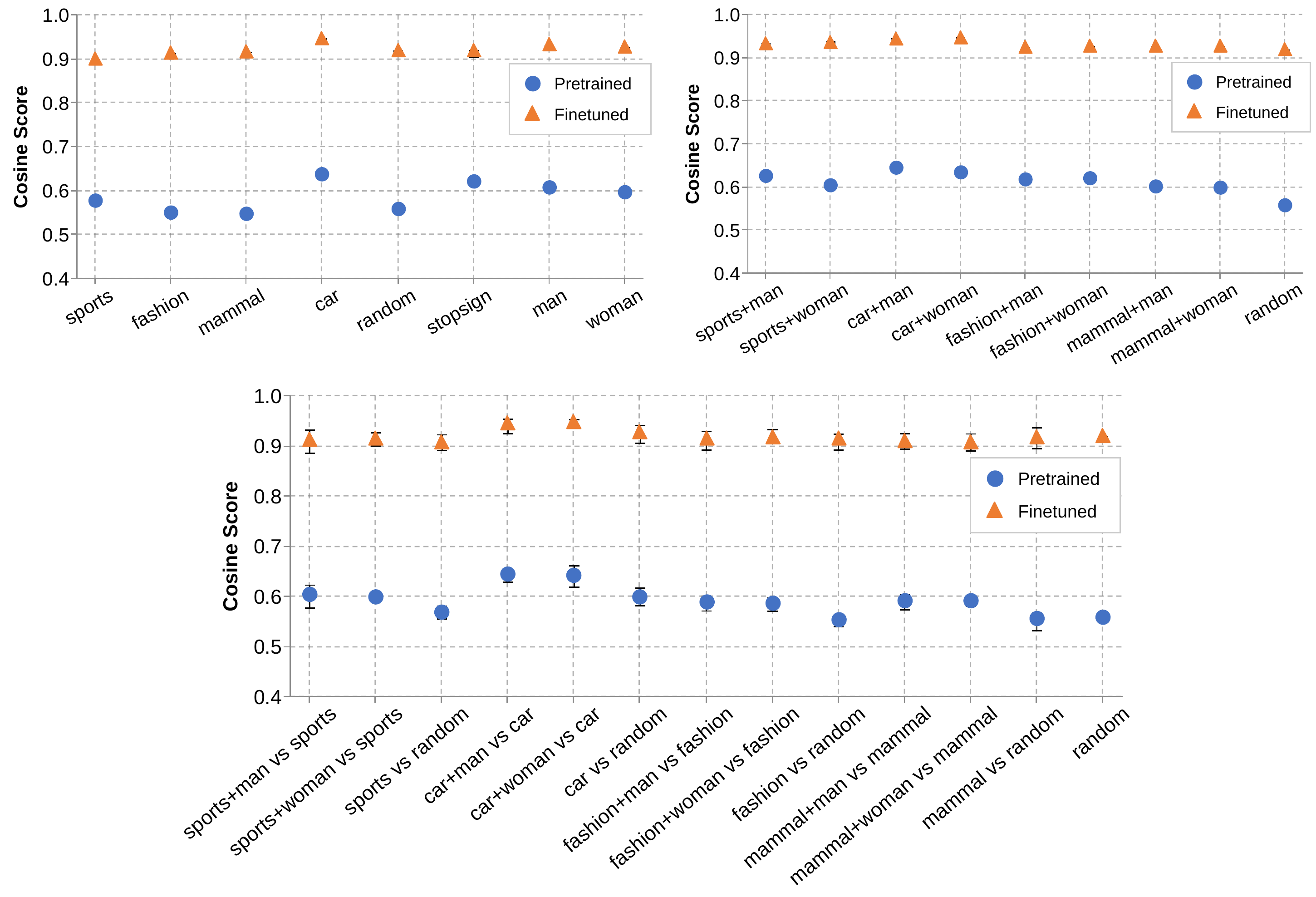} }}%
    \caption{ResNet18 finetuned on Open Images and evaluated on Open Images analysis set.}
    \label{fig:openimages_resnet18}%
\end{figure*}

\begin{figure*}[h]
    \centering {{\includegraphics[width=0.85\textwidth]{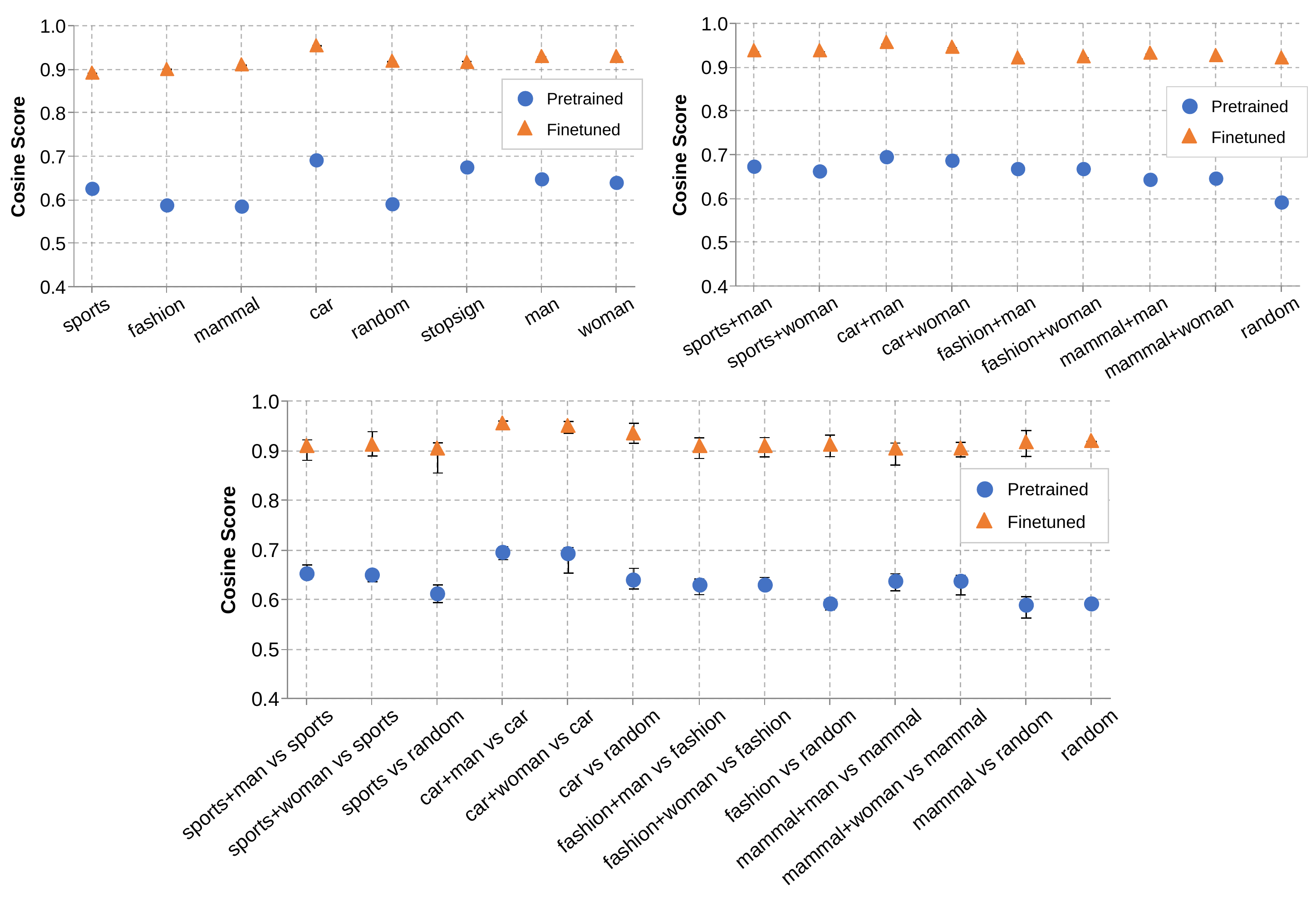} }}%
    \caption{ResNet50 finetuned on Open Images and evaluated on Open Images analysis set.}
    \label{fig:openimages_resnet50}%
\end{figure*}

\begin{figure*}[h]
    \centering {{\includegraphics[width=0.85\textwidth]{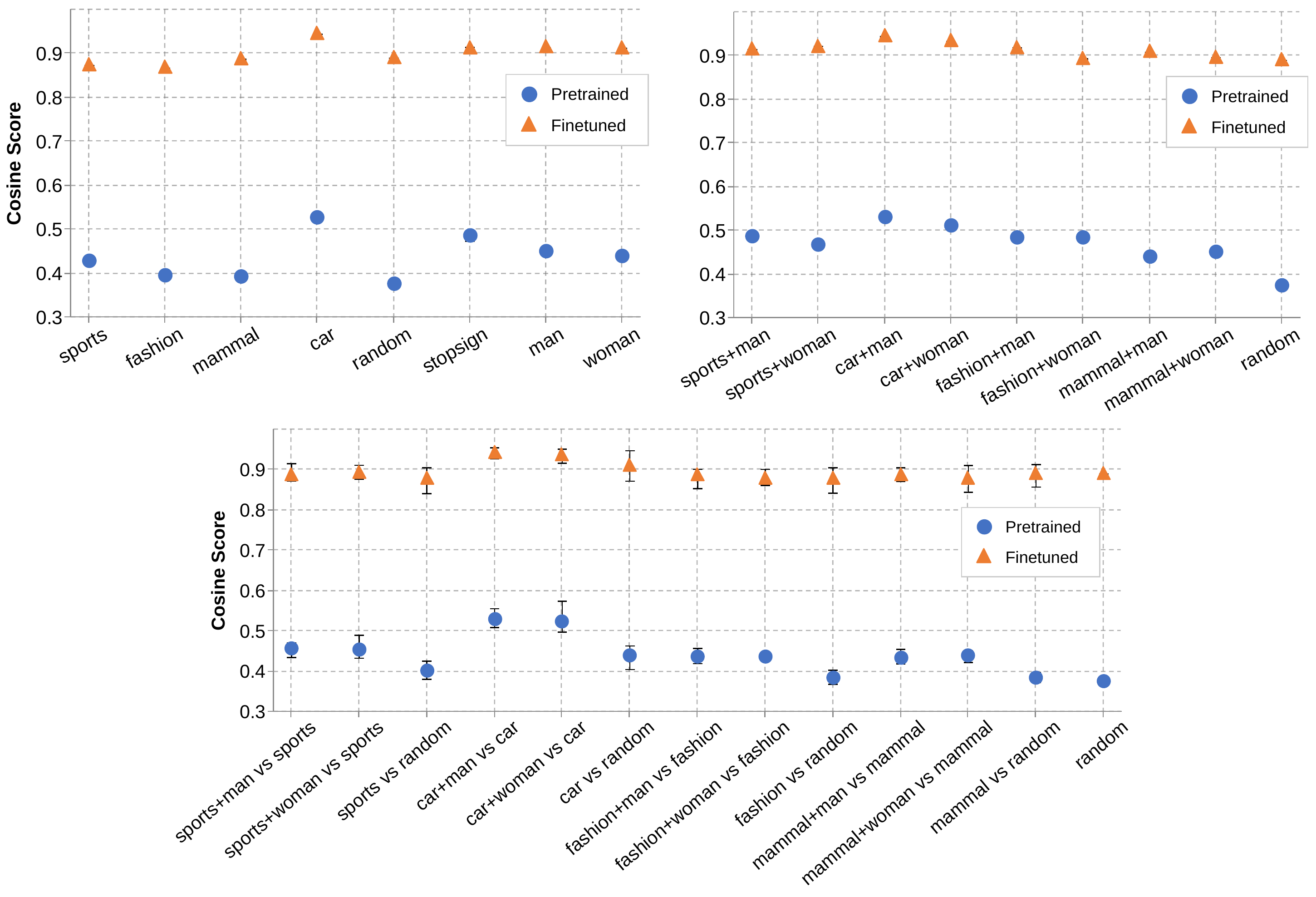} }}%
    \caption{SimCLR ResNet50 finetuned on Open Images and evaluated on Open Images analysis set.}
    \label{fig:openimages_simclr}%
\end{figure*}

We provide a complete set of results for each model finetuned on Open Images v4 and evaluated on the Open Images \emph{analysis set}. Figures $ [\ref{fig:openimages_bit}, \ref{fig:openimages_clip}, \ref{fig:openimages_moco}, \ref{fig:openimages_resnet18}, \ref{fig:openimages_resnet50},\ref{fig:openimages_simclr}]$ show the intra-class (top row) and inter-class similarities (bottom row) before and after finetuning a model on Open Images. We observe in Figure \ref{fig:openimages_bit} that BiT-M-R50x1 preserves the associations even after finetuning as can be noted by the similar trends between pretraining and finetuning similarities. This is also true for intra-class similarities for ResNet18 and SimCLR ResNet50 as can be observed in Figures \ref{fig:openimages_resnet18} and \ref{fig:openimages_simclr}. For inter-class similarities, both ResNet18 and ResNet50 acquire new biases after finetuning as can be observed by the lower correlations in Figures \ref{fig:openimages_resnet18} and \ref{fig:openimages_resnet50}. As observed in Figures $[\ref{fig:openimages_clip}, \ref{fig:openimages_moco}, \ref{fig:openimages_resnet50}]$ from the lower correlations, finetuning on a larger dataset such as Open Images can introduce new biases.

\clearpage

\end{document}